\documentclass{article}

\PassOptionsToPackage{numbers, sort}{natbib}  

\usepackage[preprint]{neurips_2020}
\usepackage{footmisc}
\usepackage[utf8]{inputenc} 
\usepackage[T1]{fontenc}    
\usepackage{hyperref}       
\usepackage{url}            
\usepackage{amsfonts}       
\usepackage{microtype}      
\usepackage{multirow}
\usepackage{tabularx}
\usepackage{adjustbox}
\usepackage{graphicx}
\usepackage{subfigure}
\usepackage{wrapfig}
\usepackage{algorithm}
\usepackage{amsmath,amsthm,amssymb}
\usepackage{bm}
\usepackage{sidecap}
\usepackage{threeparttable}
\usepackage{xcolor, color, colortbl} 

\usepackage{algorithmic}
\usepackage{booktabs}

\newcommand{\xb}{\bm{x}}
\newcommand{\ub}{\bm{u}}

\usepackage{xfrac}

\let\oldcite\cite
\renewcommand{\cite}[1]{\mbox{\oldcite{#1}}}

\title{SAMBA: Safe Model-Based \& Active Reinforcement Learning}

\author{
    Alexander I. Cowen-Rivers\thanks{The first three authors are joint first authors.} \\
    Huawei R\&D UK \\
    \texttt{alexander.cowen.rivers@huawei.com}
    \And
    Daniel Palenicek\textsuperscript{*} \\
    Huawei R\&D UK \\ Technical University Darmstadt\\
    \texttt{daniel.palenicek@stud.tu-darmstadt.de}
    \AND
    Vincent Moens\textsuperscript{*} \\
    Huawei R\&D UK \\
    \texttt{vincent.moens@huawei.com}
    \And
    Mohammed Amin ABDULLAH \\
    Huawei R\&D UK \\
    \texttt{mohammed.abdullah@huawei.com}
    \And
    Aivar Sootla \\
    Huawei R\&D UK \\
    \texttt{aivar.sootla@huawei.com}
    \And
    Jun Wang \\
    Huawei R\&D UK \\ University College London \\
    \texttt{w.j@huawei.com}
    \AND
    Haitham Bou-Ammar\thanks{Honorary position at UCL.} \\
   Huawei R\&D UK \\ University College London\\
   \texttt{haitham.ammar@huawei.com}
}

\begin{document}

\maketitle

\begin{abstract}

In this paper, we propose SAMBA, a novel framework for safe reinforcement learning that combines aspects from probabilistic modelling, information theory, and statistics. Our method builds upon PILCO to enable active exploration using novel \mbox{(semi-)metrics} for out-of-sample Gaussian process evaluation optimised through a multi-objective problem that supports conditional-value-at-risk constraints. We evaluate our algorithm on a variety of safe dynamical system benchmarks involving both low and high-dimensional state representations. Our results show orders of magnitude reductions in samples and violations compared to state-of-the-art methods. Lastly, we provide intuition as to the effectiveness of the framework by a detailed analysis of our active metrics and safety constraints.
\end{abstract}

\section{Introduction}
Reinforcement learning (RL) has seen successes in domains such as video and board games~\cite{mnih2013playing, silver2016mastering, silver2017mastering}, and control of simulated robots~\cite{ammar2014online, schulman2015trust,schulman2017proximal}. Though successful, these applications assume idealised simulators and require tens of millions of agent-environment interactions typically performed by randomly exploring policies. In real-world safety-critical applications, however, such an idealised framework of random exploration with the ability to gather samples at ease falls short, partly due to the catastrophic costs of failure and the high operating costs. Hence, if RL algorithms are to be applied in the real world, safe agents that are sample-efficient and capable of mitigating risk need to be developed. To this end, different works adopt varying safety definitions, where some are interested in safe learning, i.e., safety \emph{during} the learning process, while others focus on acquiring safe policies \emph{eventually}. Safe learning generally requires some form of safe initial policy as well as regularity assumptions on dynamics, all of which depend on which notion of safety is considered. For instance, safety defined by constraining trajectories to safe regions of the state-action space is studied in, e.g.,~\cite{akametalu2014reachability}, which assumes partially-known  control-affine dynamics with Lipschitz regularity conditions, as well as in~\cite{koller2018learning, aswani2013provably}, both of which require strong assumptions on dynamics and initial control policies to give theoretical guarantees of safety. Safety in terms of Lyapunov stability~\cite{khalil2002nonlinear} is studied in, e.g.,~\cite{chow2018lyapunov,chow2019lyapunov}~(model-free), which require a safe initial policy, and~\cite{berkenkamp2017safe}~(model-based), which requires Lipschitz dynamics and a safe initial policy. The work in~\cite{wachi2018safe}~(which builds on~\cite{turchetta2016safe}) considers deterministic dynamics and attempts to optimise expected return while not violating a pre-specified safety threshold. Several papers attempt to keep expectation constraints satisfied during learning, e.g.,~\cite{achiam2017constrained} extends ideas of~\cite{kakade2002approximately},~\cite{dalal2018safe} adds a safety layer which makes action corrections to maintain safety. When it comes to safe final policies (defined in terms of corresponding constraints), on the other hand, some works~\cite{chow2014algorithms, prashanth2014policy} considered risk-based constraints and developed model-free solvers. 

Unfortunately, most of these methods are sample-inefficient and make a large number of visits to unsafe regions. Given our interest in algorithms achieving safe final policies while reducing the number of visits to unsafe regions, 
we pursue a safe model-based framework that assumes no safe initial policies nor \emph{a priori} knowledge of the transition model. As we believe that sample efficiency is key for effective safe solvers, we choose Gaussian processes~\cite{GPbook} for our model. Of course, any such framework is hampered by the quality and assumptions of the model's hypothesis space. Aiming at alleviating these restrictions, we go further and integrate \emph{active} learning (discussed in Section~\ref{Active_learning_section}), wherein the agent influences where to query/sample from in order to generate new data so as to reduce model uncertainty, and so ultimately to learn more efficiently.  Successful application of this approach is very much predicated upon the chosen method  of quantifying potential uncertainty reduction, i.e.,  what we refer to as the \emph{(semi-)active metric}. Common (semi-)active metrics are those which identify points in the model with large entropy or large variance, or where those quantities would be most reduced in a posterior model if the given point were added to the data set~\cite{krause2007nonmyopic,krause2008near, fedorov2013theory, settles2009active}. However, our desire for safety adds a complication, since a safe learning algorithm will likely have greater uncertainty in regions where it is unsafe by virtue of not exploring those regions~\cite{deisenroth2011pilco, kamthe2017data}. Indeed our experiments (Figure~\ref{fig:igplotscdp}) support this claim.

Attacking the above challenges, we propose two novel out-of-sample \mbox{(semi-)metrics} for Gaussian processes that allow for exploration in novel areas while remaining close to training data, thus avoiding unsafe regions. To enable effective and grounded introduction of active exploration and safety constraints, we define a novel constrained bi-objective formulation of RL and provide a policy multi-gradient solver that is proven effective on a variety of safety benchmarks. In short, our contributions can be stated as follows: 1) novel constrained bi-objective formulation enabling exploration and safety consideration, 2) safety-aware active \mbox{(semi-)metrics} for exploration, and~3)~policy multi-gradient solver trading off cost minimisation, exploration maximisation, and constraint feasibility. We test our algorithm on three stochastic dynamical systems after augmenting these with safety regions and demonstrate a significant reduction in sample and cost complexities compared to the state-of-the-art.


\section{Background and notation}
\subsection{Reinforcement learning}\label{Sec:MBRL}
We consider Markov decision processes (MDPs) with continuous states and action spaces; $\mathcal{M} = \left\langle \mathcal{X}, \mathcal{U}, \mathcal{P}, c, \gamma \right\rangle$, where $\mathcal{X} \subseteq \mathbb{R}^{d_{\text{state}}}$ denotes the state space, $\mathcal{U} \subseteq \mathbb{R}^{d_{\text{act}}}$ the action space, $\mathcal{P}: \mathcal{X} \times \mathcal{U} \times \mathcal{X} \rightarrow [0, 1]$ 
is a transition density function, $c: \mathcal{X} \times \mathcal{U} \rightarrow \mathbb{R}$ is the cost function and $\gamma \in [0,1]$ is a discount factor. At each time step $t=0, \ldots, T$, the agent is in state $\xb_{t} \in \mathcal{X}$ and chooses an action $\ub_{t} \in \mathcal{U}$ transitioning it to a successor state $\xb_{t+1} \sim \mathcal{P}\left(\xb_{t+1}|\xb_{t}, \ub_{t}\right)$, and yielding a cost $c(\xb_{t}, \ub_{t})$. Given a state $\xb_t$, an action $\ub_t$ is sampled from a policy $\pi: \mathcal{X} \times {U} \rightarrow [0,1]$, where we write $\pi(\ub_{t}|\xb_{t})$ to represent the conditional density of an action. Upon subsequent interactions, the agent collects a trajectory $\bm{\tau}= [\bm{x}_{0:T}, \bm{u}_{0:T}]$, and aims to determine an optimal policy $\pi^{\star}$ by minimising total expected cost: $\pi^{\star} \in \arg\min_{\pi}\, \mathbb{E}_{\bm{\tau} \sim p_{\pi}(\bm{\tau})}[\mathcal{C}(\bm{\tau}):=\sum_{t =  0}^{T}\gamma^{t}c(\xb_t, \bm{u}_{t})]$, 
where $p_{\pi}(\bm{\tau})$ denotes the trajectory density defined as: $p_{\pi}(\bm{\tau}) = \mu_{0}(\xb_{0})\prod_{t=0}^{T-1}\mathcal{P}(\xb_{t+1}|\xb_{t}, \ub_{t})\pi(\ub_{t}|\xb_{t})$, with $\mu_{0}(\cdot)$ being an initial state distribution. \\
\underline{\textbf{Constrained MDPs:}} The above can be generalised to include various forms of constraints, often motivated by the desire to impose some form of safety measures. Examples are expectation constraints~\cite{achiam2017constrained, altman1999constrained} (which have the same form as the objective, i.e., expected discounted sum of costs), constraints on the variance of the return~\cite{prashanth2013actor}, chance constraints (a.k.a. Value-at-Risk (VaR))~\cite{chow2017risk}, and Conditional Value-at-Risk (CVaR)~\cite{chow2014algorithms, chow2017risk,prashanth2014policy}. The latter is the constraint we adopt in this paper for reasons that will be elucidated upon below. Adding constraints means we can't directly apply standard algorithms like policy gradient~\cite{sutton2018reinforcement}, and different techniques are required, e.g., via Lagrange multipliers~\cite{bertsekas1997nonlinear}, as was done in~\cite{chow2014algorithms, chow2017risk, prashanth2014policy} besides many others. Further, current methods only consider cost minimisation with no regard to exploration as we do in this paper.\\
\underline{\textbf{Model-Based Reinforcement Learning:}} Current solutions to the problem described above (constrained or unconstrained) can be split into model-free and model-based methods. Though effective, model-free algorithms are highly sample inefficient~\cite{hessel2018rainbow}. For sample-efficient solvers, we follow model-based strategies that we now detail. To reduce the number of interactions with the real environments, model-based solvers build surrogate models, $\mathcal{P}_{\text{surr}}$, to determine optimal policies. These methods, typically, run two main loops. The first gathers traces from the real environment to update $\mathcal{P}_{\text{surr}}$, while the second improves the policy using $\mathcal{P}_{\text{surr}}$~\cite{deisenroth2011pilco, hafner2019dream}. Among various candidate models, e.g.,  world models~\cite{ha2018world}, in this paper, we follow PILCO~\cite{deisenroth2011pilco} and adopt Gaussian processes (GPs) as we believe that uncertainty quantification and sample efficiency are key for real-world considerations of safety. In this construction, one places a Gaussian process prior on a latent function $f$ to map between input-output pairs. Such a prior is fully  specified by a mean, $m(\xb) = \mathbb{E}\left[f(\xb)\right]$, and a covariance function $k(\xb, \xb') =  \mathbb{E}\left[(f(\xb)-m(\xb))(f(\xb')-m(\xb'))\right]$~\cite{GPbook}. We write $f \sim \mathcal{GP}\left(m(\cdot), k(\cdot,\cdot)\right)$ to emphasize that $f$ is sampled from a GP~\cite{GPbook}. Given a data-set of input-output pairs $\{(\xb^{(i)},y^{(i)})\}^{n_{1}}_{i=1}$, corresponding, respectively, to state-action and successor state tuples, one can perform predictions on a query set of $n_{2}$ test data points $\{\bm{x}_{\star}^{(j)}\}_{j=1}^{n_{2}}$. Such a distribution is Gaussian with predictive mean-vectors and covariance matrices given by: $\bm{\mu}_{\star} = \bm{K}_{n_{2},n_{1}}\bm{A}_{n_{1},n_{1}}\bm{y}_{n_{1}}$ and $\bm{\Sigma}_{n_{2}, n_{2}} = \bm{K}_{n_{2}, n_{2}} - \bm{K}_{n_{2},n_{1}}\bm{A}_{n_{1}, n_{1}}\bm{K}_{n_{1},n_{2}}$, where $\bm{A}_{n_{1}, n_{1}} = [\bm{K}_{n_1, n_1} + \sigma_{\omega}^{2}\bm{I}]^{-1}$ with $\sigma_{\omega}$ being the noise covariance that is assumed to be Gaussian. In the above, we also defined $\bm{y}_{n_{1}}$ as a vector concatenating all training labels, $\bm{K}_{n_{1}, n_{1}} = K(\bm{X}, \bm{X}) = [k(\xb^{(i)},\xb^{(j)})]_{ij}$, $\bm{K}_{n_{1}, n_{2}}= \bm{K}_{n_{2}, n_{1}}^{\mathsf{T}} = K(\bm{X}, \bm{X}_{\star}) = [k(\xb^{(i)},\xb_*^{(j)})]_{ij}$, and $\bm{K}_{n_{2}, n_{2}}(\bm{X}_{\star},\bm{X}_{\star}) = [k(\xb_*^{(i)},\xb_*^{(j)})]_{ij}$, where $\bm{X}$ and $\bm{X}_{\star}$ are feature matrices with $\#\text{input-dim} \times n_{1}$ and $\#\text{input-dim} \times n_{2}$ sizes respectively. We executed training in GPyTorch~\cite{gardner2018gpytorch}, and used multi-output-GPs as defined in~\cite{wolff2020mogptk}. 
\subsection{Active learning in dynamical systems}\label{Active_learning_section}
In \emph{active learning}~\cite{fedorov2013theory, settles2009active}, an agent chooses points to sample/query that best improve learning or model updates. This is often performed by optimising an \emph{acquisition function}, which gives some quantification of how much a model would improve if a given data point were queried, e.g., points where the model has high entropy or where variance can be most reduced. Active learning with GPs has been studied in the static case, where points can be selected at will (see, e.g.,~\cite{krause2007nonmyopic, krause2008near}). In the context of dynamical systems, however, added complications arise as one is not always able to directly drive the system into a desired state. Recent work has attempted to resolve this problem, e.g.,  in~\cite{buisson2019actively} and~\cite{schultheis2019receding}, receding horizon optimisation is used to iteratively update a model, and in~\cite{buisson2019actively},  actions are favoured that maximise the sum of differential entropy terms at each point in the mean trajectory predicted to occur by those actions. Moreover, in~\cite{schultheis2019receding}, a sum of variance terms is optimised to improve Bayesian linear regression. Again, for computational tractability, the predicted mean of states is used as propagating state distributions in the model is difficult. Different to our paper, neither of these works deal with safety, nor do they have additional objectives to maximise/minimise avoiding a bi-objective formulation. In~\cite{jain2018learning} a GP model that is used for MPC is updated by greedily selecting points which maximise \emph{information gain}, i.e., reduction in entropy, as is done  in~\cite{krause2008near}. Only very recently, the authors in~\cite{ball2020ready} proposed an active learning approach coupled with MBRL. Similar to SAMBA, they use an adaptive convex combination of objectives, however their exploration metric is based on reward variance computed from a (finite) collection of models increasing the burden on practitioners who now need to predefine the collection of dynamics. They do not use GPs as we do, and do not consider safety. Compared to~\cite{ball2020ready}, we believe SAMBA is more flexible supporting model-learning from scratch and enabling principled exploration coupled with safety consideration. Further afield from our work, active learning has been recently studied in the context of GP time-series in~\cite{zimmer2018safe}, and for pure exploration in~\cite{shyam2018model}, which uses a finite collection of models. Our \mbox{(semi-)metrics} generalise the above to consider safe-regions and future information trade-off as we detail in Section~\ref{Sec:SafeExp}. 
\vspace{-1em}
\section{SAMBA: Framework \& solution}
In designing SAMBA, we take PILCO~\cite{deisenroth2011pilco} as a template and introduce two novel ingredients allowing for active exploration and safety. Following PILCO, SAMBA runs a main loop that gathers traces from the real environment and updates a surrogate model, $\mathcal{P}_{\text{GP}}(\cdot): \mathcal{X} \times \mathcal{U} \times \mathcal{X} \rightarrow [0,1]$, encoded by a Gaussian process. Given $\mathcal{P}_{\text{GP}}(\cdot)$, PILCO and other model-based methods~\cite{srinivas2020curl} attempt to obtain a policy that minimises total-expect cost with respect to traces, $\bm{\tau}$, sampled from the learnt model by solving $\min_{\pi} \mathbb{E}_{\bm{\tau} \sim p_{\text{surr}}(\bm{\tau})}\left[\mathcal{C}(\bm{\tau})\right]$ with $p_{\text{surr}}(\bm{\tau}) = \mu_{0}(\bm{x}_0) \prod_{t=0}^{T-1}\mathcal{P}_{\text{GP}}(\bm{x}_{t+1}|\bm{x}_{t}, \bm{u}_{t})\pi(\bm{u}_{t}|\bm{x}_{t})$. The updated policy is then used to sample new traces from the real system where the above process repeats. During this sampling process, model-based algorithms consider various metrics in acquiring transitions that reveal novel information, which can be used to improve the surrogate model's performance. PILCO, for instance, makes use of the GP uncertainty, while ensemble models~\cite{saphal2020seerl, van2020simple} explore by their aggregated uncertainties. With sufficient exploration, this allows policies obtained from surrogate-models to control real-systems. Our safety considerations mean we would prefer agents that learn well-behaving policies with minimal sampling from unsafe regions of state-action spaces; a property we achieve later by incorporating CVaR constraints as we detail in Section~\ref{Sec:ProbDef}. Requiring a reduced number of visits to unsafe regions, hence, lessens the amount of ``unsafe'' data gathered in such areas by definition. Therefore, model entropy is naturally increased in these territories and algorithms following such exploration strategies are, as a result, encouraged to sample from hazardous states. As such, a naive adaptation of entropy-based exploration can quickly become problematic by contradicting safety requirements. To circumvent these problems, we introduce two new active exploration \mbox{(semi-)metrics} in Section~\ref{Sec:SafeExp}, that assess information beyond training-data availability and consider input-output data distributions. Our \mbox{(semi-)metrics} operate under the assumption that during any model update step, ``safe'' transition data (i.e., a set of state-action-successor states sampled from safe regions) is more abundant in number when compared to ``unsafe'' triplets. Considering such a skew between distributions, our \mbox{(semi-)metrics} yield increased values on test queries close to high-density training data. Given such \mbox{(semi-)metrics}, we enable novel model-based algorithms that solve a bi-objective optimisation problem that attempts to minimise cost, while maximising active values. In other words, during this step, we not only update policies to be well-behaving in terms of the total cost but also to actively explore safe transitions that allow for improved models in successor iterations. 

Of course, the assumption of having skew towards safe regions in training data distribution is generally not true since solving the above only ensures good expected returns. 
To frequently sample safe regions, we augment cost minimisation with a safety constraint that is encoded through the CVaR of a user-defined safety cost function with respect to \emph{model} traces. Hence, SAMBA solves a bi-objective constrained optimisation problem (\S \ref{Sec:ProbDef}) aimed at minimising cost, maximising active exploration, and meeting safety constraints. 

\subsection{Bi-objective constrained MDPs}\label{Sec:ProbDef}
Given a (GP) model of an environment,  we formalise our problem as a generalisation of constrained MDPs to support bi-objective losses. We define a bi-objective MDP by a tuple $\mathcal{M}_{\text{BiO}} = \left\langle \mathcal{X}, \mathcal{U}, \mathcal{P}_{\text{GP}}, c, l, \zeta, \gamma \right\rangle$ consisting of state space $\mathcal{X} \subseteq \mathbb{R}^{d_{\text{state}}}$, action space $\mathcal{U} \subseteq \mathbb{R}^{d_{\text{act}}}$, Gaussian process transition model $\mathcal{P}_{\text{GP}}$, cost function $c: \mathcal{X} \times \mathcal{U} \rightarrow \mathbb{R}$, constraint cost function (used to encode safety) $l: \mathcal{X} \times \mathcal{U} \rightarrow \mathbb{R}$,  additional objective function  $\zeta: \mathcal{X} \times \mathcal{U} \rightarrow \mathbb{R}$, and discount factor $\gamma \in [0,1)$.\footnote{It is worth noting that the \mbox{(semi-)metrics} we develop in Section~\ref{Sec:SafeExp} map to $\mathbb{R}_{+}$ instead of $\mathbb{R}$.} In our setting, for instance, $l$ encodes the state-action's risk by measuring the distance to an unsafe region, while $\zeta$ denotes an active (semi-)metric from these described in Section~\ref{Sec:SafeExp}. To finalise the problem definition of bi-objective MDPs, we need to consider an approachable constraint to describe safety considerations. In incorporating such constraints, we are chiefly interested in those that are flexible (i.e., can support different user-designed safety criteria) and allow us to quantify events occurring in tails of cost distributions. When surveying single-objective constrained MDPs, we realise that the literature predominately focuses on expectation-type constraints~\cite{achiam2017constrained, raybenchmarking} -- a not so flexible approach restricted to being safe on average. Others, however, make use of conditional-value-at-risk~(CVaR); a coherent risk measure~\cite{chow2017risk} that provides powerful and flexible notions of safety (i.e., can support expectation, tail-distribution, or hard -- unsafe state visitation -- constraints) and quantifies tail risk in the worst $(1-\alpha)$-quantile. Formally, given a random variable $Z$, $\text{CVaR}_{\alpha}(Z)$ is defined as: $\text{CVaR}_{\alpha}(Z) = \min_{\nu} [\nu + \frac{1}{1-\alpha}\mathbb{E} [(Z - \nu)^{+}]]$, where $(Z - \nu)^{+} = \max (Z - \nu, 0)$. With such a constraint, we can write the optimisation problem of our bi-objective MDP as: 
\begin{equation}
\label{Eq:BiMDP}
    \min_{\pi} \mathbb{E}_{\bm{\tau} \sim p_{\text{surr}}(\bm{\tau})}\left[\mathcal{C}(\bm{\tau}), \zeta(\bm{\tau})\right]^{\mathsf{T}} \ \ \text{s.t.} \ \ \text{CVaR}_{\alpha}(\mathcal{L}(\bm{\tau})) \leq \xi,
\end{equation}
with $\mathcal{L}(\bm{\tau}) =  \sum_{t=0}^{T} \gamma^{t} l(\bm{x}_{t}, \bm{u}_{t})$ being total accumulated safety cost along $\bm{\tau}$, and $\xi \in \mathbb{R}_{+}$ a safety threshold. Of course, Equation~\ref{Eq:BiMDP} describes a problem not standard to reinforcement learning. In Section~\ref{Sec:SolMethod}, we devise \emph{policy multi-gradient} updates to determine $\pi$. 

\subsection{$\zeta$-functions for safe active exploration}\label{Sec:SafeExp}
In general, $\zeta$ can be any bounded objective that needs to be maximised/minimised in addition to standard cost. Here, we choose one that enables active exploration in safe state-action regions. To construct $\zeta$, we note that a feasible policy -- i.e., one abiding by CVaR constraints -- of the problem in Equation~\ref{Eq:BiMDP} samples tuples that mostly reside in safe regions. As such, the training data distribution is skewed in the sense that safe state-action pairs are more abundant than unsafe ones. Exploiting such skewness, we can indirectly encourage agents to sample safe transitions by maximising information gain \mbox{(semi-)metrics} that only grow in areas close-enough to training data. 

\paragraph{$\zeta_{\text{LOO}}$: Leave-One-Out semi-metric}
Consider a GP dynamics model, $\mathcal{P}_{\text{GP}}$, that is trained on a state-action-successor-state data set $\mathcal{D} = \{\langle \tilde{\bm{x}}^{(i)}, y^{(i)} \rangle\}_{i=1}^{n_{1}}$ with $\tilde{\bm{x}}^{(i)} = (\bm{x}^{(i)}, \bm{u}^{(i)})$.\footnote{As $\mathcal{D}$ changes at every outer iteration, we simply concatenate all data in one larger data set; see Algorithm~\ref{Algo:SAMBA}.} Such a GP induces a posterior allowing us to query predictions on $n_{2}$ test points $\tilde{\bm{x}}_{\star} = \{( \bm{x}_{\star}^{(j)}, \bm{u}_{\star}^{(j)})\}_{j=1}^{n_{2}}$. As noted in Section~\ref{Sec:MBRL}, the posterior is also Gaussian with the following mean vector and covariance matrix:\footnote{For clarity, we describe a one-dimensional scenario. We extend to multi-output GPs in our experiments.} $\bm{\mu}_{\star} = \bm{K}_{n_{2}, n_{1}}\bm{A}_{n_{1}, n_{1}}\bm{y}_{n_{1}} \ \ \text{and} \ \ \bm{\Sigma}_{n_{2}, n_{2}} = \bm{K}_{n_{2}, n_{2}} - \bm{K}_{n_{2}, n_{1}}\bm{A}_{n_{1}, n_{1}}\bm{K}_{n_{1}, n_{2}}$. Our goal is to design a measure that increases in regions with dense training-data (due to the usage of CVaR constraint) to aid agents in exploring novel yet safe tuples. To that end, we propose using an expected leave-one-out semi-metric between two Gaussian processes defined, for a one query data point $\tilde{\bm{x}}_{\star}$, as: $
    \zeta_{\text{LOO}} (\tilde{\bm{x}}_{\star}) =\mathbb{E}_{i\sim \text{Uniform}[1, n_{\text{1}}]} \left[\text{KL}\left(p(\bm{f}_{\star}|\mathcal{D}_{\neg i})||p\left(\bm{f}_{\star}|\mathcal{D}\right)\right)\right]$ \text{with $\mathcal{D}_{\neg i}$ being $\mathcal{D}$ with point $i$ left-out.} Importantly, such a measure will only grow in regions which are close-enough to sampled training data, as posterior mean and covariance of $p(\bm{f}_{\star} | \mathcal{D}_{\neg i})$ shift by a factor that scales linearly and quadratically, respectively, with the total covariance between $\tilde{\bm{x}}_{\star}$ and $\bm{X}_{\neg i}$ where $\bm{X}_{\neg i}$ denotes a feature matrix with the $i^{th}$ row removed.\footnote{Though intuitive, we provide a formal treatment of the reason behind such growth properties in the appendix.} In other words, such a semi-metric fulfils our requirement in the sense that if a test query is distant (in distribution) from all training input data, it will achieve low $\zeta_{\text{LOO}}$ score. Though appealing, computing a full-set of $\zeta_{\text{LOO}}$ can be highly computationally expensive, of the order of $\mathcal{O}(n_{1}^{4})$ -- computing $\bm{A}_{n_{1\neg i}, n_{1\neg i}}$ requires $\mathcal{O}(n_{1}^{3})$ and this has to be repeated $n_{1}$ times. A major source contributing to this expense, well-known in GP literature, is related to the need to invert covariance matrices. Rather than following variational approximations (which constitute an interesting direction), we prioritise sample-efficiency and focus on exact GPs. To this end, we exploit the already computed $\bm{A}_{n_{1}, n_{1}}$ during the model-learning step and make-use of the matrix inversion lemma~\cite{petersen2008matrix} to recursively update the mean and covariances of $p(\bm{f}_{\star}| \mathcal{D}_{\neg i})$ for all $i$ (see appendix): $  \bm{\mu}_{\star}^{(i)} = \bm{\mu}_{\star} - \bm{K}_{n_{2}, n_{1}} \frac{\bm{a}^{\mathsf{T}}_{i} \bm{a}_{i}}{a_{i,i}} \bm{y}_{n_{1}} \ \ \text{and} \ \ \bm{\Sigma}_{n_{2}, n_{2}}^{(i)} = \bm{\Sigma}_{n_{2}, n_{2}} + \bm{K}_{n_{2}, n_{1}}\frac{\bm{a}^{\mathsf{T}}_{i} \bm{a}_{i}}{a_{i,i}} \bm{K}_{n_{1}, n_{2}}$, with $\bm{a}_{i}$ being the $i^{th}$ row of $\bm{A}_{n_{1},n_{1}}$. Hence, updating the inverse covariance matrix only requires computing and adding the outer product of the $i^{\text{th}}$ row $\bm{A}_{n_{1}, n_{1}}$, divided by the $i^{\text{th}}$ diagonal element. This, in turn, reduces complexity from $\mathcal{O}(n_{1}^{4})$ to $\mathcal{O}(n_{1}^{3})$.
\paragraph{$\zeta_{\text{Bootstrap}}$: Bootstrapped symmetric metric}
We also experimented with another metric that quantifies posterior sensitivity to bi-partitions of the data as measured by symmetric KL-divergence,\footnote{A symmetric KL-divergence between two distributions p, and q is defined as: $\sfrac{(\mathrm{KL}(p\|q)+\mathrm{KL}(q\|p))}{2}$.} averaged over possible bi-partitions: $\zeta_{\text{Bootstrap}}(\tilde{\bm{x}}_{\star}) =\mathbb{E}_{\langle \mathcal{D}_{1},\mathcal{D}_{2}\rangle}[\text{KL}_{\text{sym}}(p(\bm{f}_{\star}|\mathcal{D}_{1}) ||p(\bm{f}_{\star}|\mathcal{D}_{2}))]$, where $ \langle \mathcal{D}_{1}, \mathcal{D}_{2} \rangle$ is a random bi-partition of the data $\mathcal{D}$. In practice, we randomly split the data in half, and do this $K$ times (where $K$ is a tuneable hyper-parameter) to get a collection of $K$ bi-partitions. We then average over that collection.  Similar to $\zeta_{\text{LOO}}$, $\zeta_{\text{Bootstrap}}$ also assigns low importance to query points far from the training inputs, and hence, can be useful for safe-decision making. In our experiments, $\zeta_{\text{LOO}}$ provided better-behaving exploration strategy, see Section~\ref{Sec:Exps}.
\paragraph{Transforming $\zeta_{\cdot}(\tilde{\bm{x}}_{\star})$ to $\zeta_{\cdot}(\bm{\tau})$}
Both introduced functions are defined in terms of query test points $\tilde{\bm{x}}_{\star}$. To incorporate in Equation~\ref{Eq:BiMDP}, we define trajectory-based expected total information gain as $\bm{\zeta}_{\text{LOO}} (\bm{\tau}) = \sum_{t=0}^{T} \gamma^{t} \zeta_{\text{LOO}} (\langle \bm{x}_{t},\bm{u}_{t}\rangle) \ \ \text{and} \ \  \bm{\zeta}_{\text{Bootstrap}} (\bm{\tau}) = \sum_{t=0}^{T} \gamma^{t} \zeta_{\text{Bootstrap}} (\langle \bm{x}_{t},\bm{u}_{t}\rangle)$. Interestingly, this characterisation trades off long-term versus short-term information gain similar to how cost trades-off optimal greedy actions versus long-term decisions. In other words, it is not necessarily optimal to seek an action that maximises immediate information gain since such a transition can ultimately drive the agent to unsafe states (i.e., ones that exhibit low $\zeta_{\cdot}(\tilde{\bm{x}})$ values). In fact, such horizon-based definitions have also recently been shown to improve modelling of dynamical systems~\cite{buisson2019actively, shyam2018model}. Of course, our problem is different in the sense that we seek safe policies in a safe decision-making framework, and thus require safely exploring \mbox{(semi-)metrics}. 
\vspace{-1em}
\subsection{Solution method}
\label{Sec:SolMethod}
We now provide a solver to the problem in Equation~\ref{Eq:BiMDP}. We operate using $\bm{\zeta}_{\text{LOO}}(\bm{\tau})$ and note that our derivations can exactly be repeated for $\bm{\zeta}_{\text{Bootstrap}}(\bm{\tau})$. Since we maximise exploration, we use $-\bm{\zeta}_{\text{LOO}}(\bm{\tau})$ in the minimisation problem in Equation~\ref{Eq:BiMDP}. Effectively, we need to overcome two hurdles for an implementable algorithm. The first, relates to the bi-objective nature of our problem, while the second is concerned with the CVaR constraint that requires a Lagrangian-type solution. \\ \\
\underline{\textbf{From Bi- to Single objectives:}} We transform the bi-objective problem into a single objective one through a linear combination of $\mathcal{C}(\bm{\tau})$ and $\zeta_{\text{LOO}}(\bm{\tau})$. This relaxed yet constrained version is given by $\min_{\pi} \lambda_{\pi} \mathbb{E}_{\bm{\tau} \sim p_{\text{surr}}(\bm{\tau})}[\mathcal{C}(\bm{\tau})] - (1-\lambda_{\pi})\mathbb{E}_{\bm{\tau} \sim p_{\text{surr}}(\bm{\tau})}[\bm{\zeta}_{\text{LOO}}(\bm{\tau})] \ \ \text{s.t.} \ \ \text{CVaR}_{\alpha}(\mathcal{L}(\bm{\tau})) \leq \xi$,\footnote{Please note that the negative sign in the linear combination is due to the fact that we used $-\bm{\zeta}_{\text{LOO}}(\bm{\tau})$.} where $\lambda_{\pi}$ is a policy dependent weighting. The choice of $\lambda_{\pi}$, however, can be difficult as not any arbitrary combination is acceptable as it has to ultimately yield a Pareto-efficient solution.
Fortunately, the authors in~\cite{sener2018multi} have demonstrated that a theoretically-grounded choice for such a weighting in a stochastic multi-objective problem is one that produces a common descent direction that points opposite to the minimum-norm vector in the convex hull of the gradients, i.e., one solving: $\lambda_{\pi}^{\star} = \arg\min_{\lambda_{\pi}} || \lambda_{\pi} \nabla_{\pi} \mathbb{E}_{\bm{\tau}}[\mathcal{C}(\bm{\tau})] - (1 - \lambda_{\pi})\nabla_{\pi} \mathbb{E}_{\bm{\tau}}[\bm{\zeta}_{\text{LOO}}(\bm{\tau})]||_{2}^{2}$. Luckily,  solving for $\lambda_{\pi^{\star}}$ is straight-forward and can be encoded using a rule-based strategy that compares gradient norms, see appendix for further details. \\
\underline{\textbf{From Constrained to Unconstrained Objectives:}} We write an unconstrained problem using a Lagrange multiplier $\lambda_{\text{CVaR}}$:
$\min_{\pi}  \lambda_{\pi}^{\star} \mathbb{E}_{\bm{\tau} \sim p_{\text{surr}}(\bm{\tau})}[\mathcal{C}(\bm{\tau})] - (1 - \lambda_{\pi}^{\star}) \mathbb{E}_{\bm{\tau} \sim p_{\text{surr}}(\bm{\tau})}[\bm{\zeta}_{\text{LOO}}(\bm{\tau})] + \lambda_{\text{CVaR}} [\text{CVaR}_{\alpha}(\mathcal{L}(\bm{\tau})) - \xi]$. Due to non-convexity of the problem, we cannot assume strong duality holds, so in our experiments, we schedule $\lambda_{\text{CVaR}}$ proportional to gradients  using a technique similar to that in~\cite{schulman2017proximal} that has proven effective.\footnote{Note that a primal dual-method as  in~\cite{chow2015risk} is not applicable due to non-convexity. In the future, we plan to study approaches from~\cite{goh2001nonlinear} to ease determining $\lambda_{\text{CVaR}}$.} To solve the above optimisation problem, we first fix $\nu$ and perform a policy gradient step in $\pi$.\footnote{We resorted to policy gradients for two reason: 1) cost functions are not necessarily differentiable, and 2) better experimental behaviour when compared to model back-prop especially on OpenAI's safety gym tasks.} To minimise the variance in the gradient estimator of $\mathcal{C}(\bm{\tau})$ and $\bm{\zeta}_{\text{LOO}}(\bm{\tau})$, we build two neural network critics that we use as baselines. The first attempts to model the value of the standard cost, while the second learns information gain values. For the CVaR's gradient, we simply apply policy gradients. As CVaR is non-Markovian, it is difficult to estimate its separate critic. In our experiments, a heuristic where discounted safety losses as unbiased baselines was used and proved effective. 
In short, our main update equations when using a policy parameterised by a neural network with parameters $\bm{\theta}$ can be written as: 
\small
\begin{equation}
\label{Eq:Update}
\begin{aligned}
    \bm{\theta}^{[j][k+1]} = \bm{\theta}^{[j][k]} - \eta_{k} \Big(\mathbb{E}_{\bm{\tau}}\Big[\sum_{t \geq 1} &\nabla_{\bm{\theta}}\log \pi_{\bm{\theta}}(\bm{u}_{t}|\bm{x}_{t})\Big(\lambda_{\pi}^{\star} (Q_{\mathcal{C}}(\bm{x}_{t},\bm{u}_{t}) - V_{\mathcal{C}}^{\phi_{1}}(\bm{x}_{t})) \\
    & - (1 -\lambda_{\pi}^{\star})(Q_{\bm{\zeta}}(\bm{x}_{t},\bm{u}_{t}) - V_{\bm{\zeta}}^{\phi_{2}}(\bm{x}_{t}))\Big)\Big]
    + \lambda_{\text{CVaR}}\nabla_{\bm{\theta}}\text{CVaR}_{\alpha}(\mathcal{L}(\bm{\tau}))\Big)
\end{aligned}
\end{equation}
\normalsize
where $\eta_{k}$ is a learning rate, and $V_{\mathcal{C}}^{\phi_{1}}(\bm{x}_{t})$, $V_{\bm{\zeta}}^{\phi_{2}}(\bm{x}_{t})$ are neural network critics with parameters $\phi_{1}$ and $\phi_{2}$. We present the main steps in Algorithm~\ref{Algo:SAMBA} and more details in the appendix. 
\begin{algorithm}
\caption{SAMBA: Safe Model-Based \& Active Reinforcement Learning}
\label{Algo:SAMBA}
\begin{algorithmic}[1]
\STATE \textbf{Inputs:} $\lambda_{\text{CVaR}}$ initialisation, initial random policy $\pi_1$, $\mathcal{D}_{0}=\{\}$, $\alpha$-quantile, safety threshold $\xi$
\STATE \textbf{for} $j=1:\#\text{env-iterations}$ \textbf{do:}
\STATE \hspace{2em} Sample traces from the real environment using $\pi_{j}$, concatenate all data and update $\mathcal{P}_{\text{GP}}^{[j]}(\cdot)$
\STATE \hspace{2em} \textbf{for} $k=1:\#\text{control-iterations}$ \textbf{do:}
\STATE \hspace{3em} Sample traces from $\mathcal{P}^{[j]}_{\text{GP}}(\cdot)$ and compute $\bm{\zeta}^{[j][k]}_{\text{LOO}}(\bm{\tau})$ (\textsection~\ref{Sec:SafeExp})
\STATE \hspace{3em} Solve for \small $\lambda_{\pi}^{[j][k],\star} = \arg\min_{\lambda_{\pi}} || \lambda_{\pi} \nabla_{\bm{\theta}} \mathbb{E}_{\bm{\tau}}[\mathcal{C}(\bm{\tau})] - (1 - \lambda_{\pi})\nabla_{\bm{\theta}} \mathbb{E}_{\bm{\tau}}[\bm{\zeta}_{\text{LOO}}(\bm{\tau})]||_{2}^{2}$ \normalsize
\STATE \hspace{3em} Update $\phi^{[j][k]}_{1}$ and $\phi^{[j][k]}_{2}$ with $\mathbb{E}_{\bm{\tau}}[ \mathcal{C}(\bm{\tau})]$ and $\mathbb{E}_{\bm{\tau}}[ \bm{\zeta}_{\text{LOO}}(\bm{\tau})]$ as targets, and $ \bm{\theta}^{[j][k]}$ (\textsection~\ref{Sec:SolMethod})
\STATE \hspace{2em} Set $\pi_{j+1} = \pi_{\#\text{control-iterations}}$
\STATE \textbf{Output:} Policy $\pi_{\#\text{env-iterations}}$
\end{algorithmic}
\end{algorithm}
\section{Experiments}\label{Sec:Exps}
\begin{figure}[h!]
\centering
\includegraphics[width=1.\linewidth]{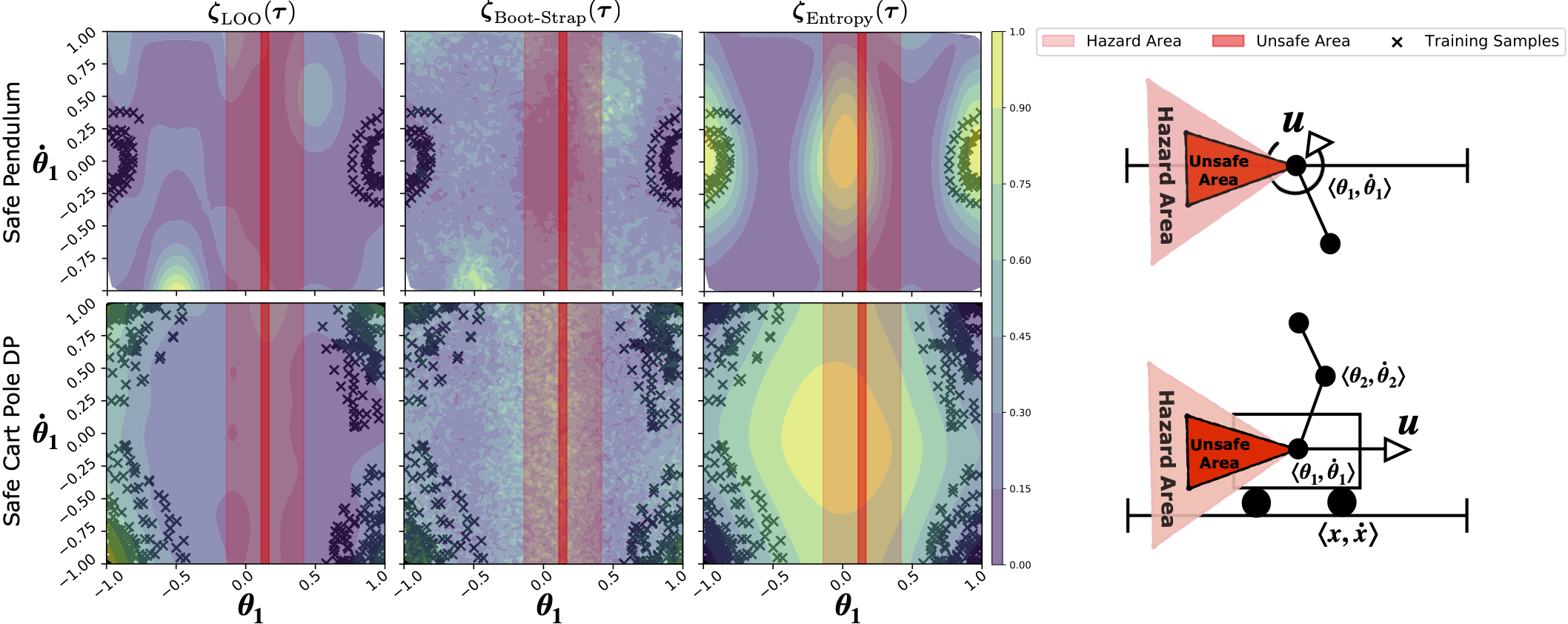}
\vspace{-2em}
\caption{
    Comparing active learning metric values across the entire state space. Note, policies used within this framework are encouraged to visit regions with higher (yellow) values.
    The dynamics model for Safe Pendulum was trained on $100$ samples and for Safe Cart Pole Double Pendulum on $500$ samples respectively. More plots for different numbers of samples can be found in the appendix.
}
\label{fig:igplotscdp}
\end{figure}

We assess SAMBA in terms of both \emph{safe learning} (train) and \emph{safe final policies} (test) on three dynamical systems, two of which are adaptations of standard dynamical systems for MBRL (Safe Pendulum and Safe Cart-Pole Double Pendulum), while the third (Fetch Robot -- optimally control end-effector to reach a 3D goal) we adapt from OpenAI's robotics environments~\cite{brockman2016openai}. 
In each of these tasks, we define unsafe regions as areas in state spaces and design the safety loss (i.e., $\mathcal{L}(\bm{\tau})$) to correspond to the (linearly proportional) distance between the end-effector's position (when in the hazard region) to the centre of the unsafe region. SAMBA implemented a more stable proximal update of Equation~\ref{Eq:Update} following a similar method to~\cite{schulman2017proximal}. We compare against algorithms from both model-free and model-based literature. Model-free comparisons against TRPO~\cite{schulman2015trust}, PPO~\cite{schulman2017proximal}, CPO~\cite{achiam2017constrained}, STRPO (safety-constrained TRPO)~\cite{raybenchmarking} and SPPO~\cite{raybenchmarking} (safety-constrained PPO) enable us to determine if SAMBA improves upon the following: sample complexities during training;  total violations~(TV), that is, the total number of timesteps spent inside the unsafe region;  total accumulated safety~cost~(TC).\footnote{Note, we report safe learning process TC, which is the total incurred safety cost throughout all training environment interactions, and safe evaluation TC and TV, which similarly is the total incurred safety cost during evaluation, and the total violations from the timesteps spent inside the unsafe region during evaluation.\label{fnlabel}} Comparison with model-based solvers (e.g., PlaNet~\cite{hafner2018learning}, (P)PILCO~\cite{deisenroth2011pilco}) sheds light on the importance of our active exploration metrics. It is important to note that when implementing PILCO, we preferred a flexible solution that does not assume moment-matching and specific radial-basis function controllers. Hence, we adapted PILCO to support proximal policy updates, referred to as PPILCO in our experiments, and similarly, SPPILCO (safety-constrained PPILCO), which also proved effective. As SAMBA introduces exploration components to standard model-based learners, we analysed these independently before combining them and reporting  TV and TC~\footref{fnlabel} (see Table~\ref{tab:modelresults}).
All policies are represented by two-hidden-layer (32 units each) neural networks with $tanh$ non-linearities. Each requires under 12 hours of training on an NVIDIA GeForce RTX 2080 Ti which has a power consumption of 225W, yielding an estimated training cost of \pounds~$0.1$ per model, per seed.
Due to space constraints, all hyper-parameters to reproduce our results can be found in the appendix. \\
\underline{\textbf{\mbox{(Semi-)metrics} Component:}} Evaluating our (semi-metrics), we conducted an analysis that reports a 2D projection view of the state space at various intervals in the data-collection process.
We compare $\zeta_{\text{LOO}}(\bm{\tau})$ and $\zeta_{\text{Bootstrap}}(\bm{\tau})$ against an entropy-based exploration metric and report the results on two systems in Figure~\ref{fig:igplotscdp}. It is clear that both our \mbox{(semi-)metrics} encourage agents to explore safe regions in the state space as opposed to entropy that mostly grows in unsafe regions. Similar results are demonstrated with the Fetch Reach robot task (in the appendix). It is also worth noting that due to the high-dimensional nature of the tasks, visual analysis can only give indications. Still, empirical analysis supports our claim and performance improvements are clear; see Table~\ref{tab:modelresults}.
\\
\underline{\textbf{Learning and Evaluation:}} Having concluded that our \mbox{(semi-)metrics} indeed provide informative signals to the agent for safe exploration, we then conducted learning and evaluation experiments comparing SAMBA against state-of-the-art methods. Results reported in Table~\ref{tab:modelresults} demonstrate that SAMBA reduces the amount of  training TC,~\footref{fnlabel} and samples by orders of magnitude compared to others. Interestingly, during safe evaluation (deploying learnt policy and evaluating performance), we see SAMBA's safety performance competitive with (if not significantly better than) policies trained for safety in terms of TC and TV.~\footref{fnlabel} Such results are interesting as SAMBA was never explicitly designed to specifically minimise test TV,~\footref{fnlabel} but it was still able to acquire significant reductions. Of course, one might argue that these results do not convey a safe final policy, as violations are still non-zero. We remind, however, that we define safety in terms of CVaR constraints, which do not totally prohibit violations, rather, limit the average cost of excess violations beyond a (user-defined) risk level $\alpha$. Indeed, as mentioned above, it is not possible to guarantee total safety without strong assumptions on dynamics and/or an initial safe policy (and of course, none of the algorithms in  Table~\ref{tab:modelresults} have zero violations). 

\begin{table}[h!]
\caption{
    Safe learning and safe evaluation for Safe Pendulum, Safe Cart Pole Double Pendulum, and Safe Fetch Reach. We averaged each policy's safe evaluation over three random seeds, and collected 10000 evaluation samples per seed.~\footref{fnlabel}
}
\resizebox{\columnwidth}{!}{
\begin{tabular}{@{}lcccccccccccc@{}}
\toprule
\textbf{Learner} & \multicolumn{4}{c}{\textbf{Safe Pendulum}} & \multicolumn{4}{c}{\textbf{Safe Cart Pole Double Pendulum}} & \multicolumn{4}{c}{\textbf{Safe Fetch Reach}} \\ \cmidrule(r){2-5} \cmidrule(r){6-9} \cmidrule(r){10-13} 
 & \multicolumn{2}{c}{Learning} & \multicolumn{2}{c}{Evaluation} & \multicolumn{2}{c}{Learning} & \multicolumn{2}{c}{Evaluation} & \multicolumn{2}{c}{Learning} & \multicolumn{2}{c}{Evaluation} \\
\cmidrule(r){2-3} \cmidrule(r){4-5} \cmidrule(r){6-7} \cmidrule(r){8-9} \cmidrule(r){10-11} \cmidrule(r){12-13}
                       & Samples & TC & TV & TC & Samples & TC & TV & TC & Samples & TC            & TV              & TC                \\ \midrule
\textcolor{blue}{TRPO}                  & 1000               & 110          & 4.3              & 4.9              & 1000               & 310          & 10             & 23             & 1000              & 790          & 1.8            & 2.7             \\
\textcolor{blue}{PPO}                    & 1000               & 81           & 3.6              & 5              & 1000               & 140          & 5.5              & 35             & 1000            & 670          & 1.6            & 3.1             \\
\textcolor{orange}{PlanNet}                & 100                & 56           & 4.5              & 5.2              & 40                 & 2.6          & 7.7              & 29             & 100              & 110          & 2.3            & 2.9             \\
\textcolor{orange}{PPILCO}                 & 2                  & 0.04        & 2.8              & 4.5              & 6                  & 0.097        & 7.1              & 33             & 5              & 0.49          & 1.9            & 3            \\
\textcolor{orange}{PlaNet w RS}            & 100                & 51           & 3.5              & 4.2              & 40                 & 1.3          & 1.6              & 2.5              & -                 & -             & -               & -                 \\
\textcolor{blue}{CPO}                    & 1000               & 58           & 4.7              & 4.4              & 1000               & 95           & 1.1              & 5              & 1000              & 160          & 1.9            & 2.7             \\
\textcolor{blue}{STRPO}                  & 1000               & 38           & 2.0              & 2.1              & 1000               & 83           & 2.1              & 2.7              & 1000              & 170          & 1.7            & 2.1             \\
\textcolor{blue}{SPPO}                   & 1000               & 65           & 1.7              & 2.4              & 1000               & 68           & 1              & 1.8              & 1000              & 290          & 1.5            & 2             \\
\textcolor{orange}{SPPILCO}                & 1.8                & 0.02         & 1.7              & 2.1              & 6                  & 0.062        & 1.2              & 1.6              & 5              & 0.37          & 1.4            & 2.2             \\
\textbf{\textcolor{orange}{SAMBA}}                  & \textbf{1.6}                & \textbf{0.01}         & \textbf{1.5}              & \textbf{2.0}              & \textbf{5}                  & \textbf{0.054}        & \textbf{0.85}             & \textbf{1.4}& \textbf{5}               & \textbf{0.23}          & \textbf{1.2} & \textbf{1.9}             \\
\bottomrule
\end{tabular}
}
\begin{tablenotes}
  \scriptsize
  \item The data is scaled by $10^3$. Note: PlaNet w RS on Safe Fetch Reach diverged during training.
\end{tablenotes}
\label{tab:modelresults}
\end{table}

To evaluate risk performance, we conducted an in-depth study evaluating all methods on the two most challenging environments, Safe Cart Pole Double Pendulum and Safe Fetch Reach, using cost limits $\xi=0.25$ and $\xi=0.75$. Table~\ref{tab:constraint_sat} demonstrates that SAMBA achieves lower safety cost quartiles and lower expected safety cost.
Therefore, we conclude that, indeed, SAMBA produces safe-final policies in terms of its objective in Equation~\ref{Eq:BiMDP}.
\begin{table}[h!]
\caption{
    Constraint satisfaction ($\color{green} \bm{\checkmark}$) or constraint violation ($\color{red} \bm{\times}$) on Safe Cart Pole Double Pendulum and Safe Fetch of model-based (\textcolor{orange}{orange}) and model-free (\textcolor{blue}{blue}) solvers. We evaluated expectation constraints (Exp.) and CVaR.
    Results show SAMBA satisfied safety constraints, outperforming others in different quartiles. There is also a clear trade-off between returns and safety for all algorithms. Interestingly, SAMBA is safer yet acquiring acceptable expected returns -- close to SPPO for instance.
}
\resizebox{\columnwidth}{!}{
\begin{tabular}{@{}lcccccccccccc@{}}
\toprule
\textbf{Learner} 
& \multicolumn{6}{c}{\textbf{Safe Cart Pole Double Pendulum}} & \multicolumn{6}{c}{\textbf{Safe Fetch Reach}} \\  \cmidrule(r){2-7} \cmidrule(r){8-13}
& \multicolumn{3}{c}{Quartile} & \multicolumn{2}{c}{Constraint} & \multicolumn{1}{c}{\multirow{2}{*}{}}
& \multicolumn{3}{c}{Quartile} & \multicolumn{2}{c}{Constraint} & \multicolumn{1}{c}{\multirow{2}{*}{}} \\ 
  \cmidrule(r){2-4} \cmidrule(r){5-6} \cmidrule(r){8-10} \cmidrule(r){11-12}
& \multicolumn{1}{c}{$0.25$} & \multicolumn{1}{c}{$0.5$} & \multicolumn{1}{c}{$0.75$}  & \multicolumn{1}{c}{Exp.} & \multicolumn{1}{c}{CVaR$_{\alpha}$} & \multicolumn{1}{c}{$\mathbb{E}_{\bm{\tau}}[ \mathcal{C}(\bm{\tau})]$}
& \multicolumn{1}{c}{$0.25$} & \multicolumn{1}{c}{$0.5$} & \multicolumn{1}{c}{$0.75$} & \multicolumn{1}{c}{Exp.} & \multicolumn{1}{c}{CVaR$_{\alpha}$}  & \multicolumn{1}{c}{$\mathbb{E}_{\bm{\tau}}[ \mathcal{C}(\bm{\tau})]$} \\
\midrule
\textcolor{blue}{PPO}              & \multicolumn{1}{c}{5.0}         & \multicolumn{1}{c}{5.7}        & \multicolumn{1}{c}{9.4}         & \multicolumn{1}{c}{$ \color{red} \bm{\times} $ }     & \multicolumn{1}{c}{$ \color{red} \bm{\times} $ }            & \multicolumn{1}{c}{\textbf{-19}}                        & 0.95 & 1.25 &  1.88  & $ \color{red} \bm{\times} $ &  $ \color{red} \bm{\times} $   & \textbf{-0.26}                     \\ 
\textcolor{orange}{PPILCO} & 5.0   &  5.8   &  8.6                                                     &             $ \color{red} \bm{\times} $              &                    $ \color{red} \bm{\times} $              &   -21                                         &    0.91  & 1.22  &  1.45  &  $ \color{red} \bm{\times} $  &  $ \color{red} \bm{\times} $  & -0.36    \\
\textcolor{orange}{PlaNet}   & 5.2       &     6.0                         &      8.8                       &                              $ \color{red} \bm{\times} $                          &    $ \color{red} \bm{\times} $    &             -21              &                                             0.64 & 1.09 & 1.88 &  $ \color{red} \bm{\times} $ &  $ \color{red} \bm{\times} $  & -0.39                                             \\
\textcolor{blue}{TRPO}   & 5.1         &       5.9                       &   7.0                          &                               $ \color{red} \bm{\times} $                        &               $ \color{red} \bm{\times} $                   &               -22                              & 0.69  & 0.79  & 0.92  &  $ \color{red} \bm{\times} $  &  $ \color{red} \bm{\times} $ & -0.58                                        \\
\textcolor{blue}{CPO}              &          0.41                    &        0.47                    &     0.96                         &             $ \color{red} \bm{\times} $              &           $ \color{red} \bm{\times} $                       &                             -23                & 0.81  & 0.92  & 1.02  &  $ \color{red} \bm{\times} $  &  $ \color{red} \bm{\times} $   &  -0.43                                          \\
\textcolor{orange}{PlaNet w RS}      &    0.83                           &         0.94                      &  1.11                             &              $ \color{red} \bm{\times} $             &         $ \color{red} \bm{\times} $                         &                              -24               &                -              &               -              &             -                 &            -              &                  -               &                          -                   \\
\textcolor{orange}{SPPILCO}          &         0.21                     &            0.28                 &      0.33                        &             $ \color{red} \bm{\times} $              &          $ \color{red} \bm{\times} $                        &                       -25                      & 0.45  & 0.77  & 1.13  &  $ \color{green} \bm{\checkmark} $ &  $ \color{red} \bm{\times} $ & -0.57                                           \\
\textcolor{blue}{STRPO}            &          0.22                    &            0.27                 &      0.32                         &            $ \color{green} \bm{\checkmark} $               &                        $ \color{red} \bm{\times} $         &                       -28                     & 0.14  & 0.46  & 0.86  & $ \color{green} \bm{\checkmark} $  &  $ \color{red} \bm{\times} $   & -0.98                                          \\
\textcolor{blue}{SPPO}             &             0.19                 &        0.24                     &       0.29                       &            $ \color{green} \bm{\checkmark} $               &             $ \color{red} \bm{\times} $                  &                            -27                 & 0.27  & 0.44  & 0.71  & $ \color{green} \bm{\checkmark} $  & $ \color{green} \bm{\checkmark} $     & -1.51                                         \\
\textbf{\textcolor{orange}{SAMBA}}            &          \textbf{0.15}                    &              \textbf{0.21}               &       \textbf{0.24}                      &                $ \color{green} \bm{\checkmark} $          &       $ \color{green} \bm{\checkmark} $                          &                          -27               & \textbf{0.00} & \textbf{0.05} & \textbf{0.19} & $ \color{green} \bm{\checkmark} $ & $ \color{green} \bm{\checkmark} $ & -2.27                                           \\
\bottomrule
\end{tabular}
}
\begin{tablenotes}
  \scriptsize
  \item Note: PlaNet w RS on Safe Fetch Reach diverged during training.
\end{tablenotes}
\label{tab:constraint_sat}
\end{table}

\vspace{-1em}
\section{Conclusion and future work}
We proposed SAMBA, a safe and active model-based learner that makes use of GP models and solves a bi-objective constraint problem to trade-off cost minimisation, exploration, and safety. We evaluated our method on three benchmarks, including ones from Open AI's safety gym and demonstrated significant reduction in training cost and sample complexity, as well as safe-final policies in terms of CVaR constraints compared to the state-of-the-art. 

In future, we plan to generalise to variational GPs~\cite{damianou2011variational}, and to apply our method in real-world robotics. Additionally, we want to study the theoretical guarantees of SAMBA to demonstrate convergence to well-behaving, exploratory, and safe policies.

\section*{Broader Impact}
Not applicable.

\bibliography{ref}
\bibliographystyle{plain}
\newpage
\appendix
\section{Algorithm}
In this section we provide a more detailed description of Algorithm~\ref{Algo:SAMBA}. We shall use the following short-hand for the advantage function: $A^{\phi_1}_{\mathcal{C}}(\bm{x},\bm{u}) := Q_{\mathcal{C}}(\bm{x},\bm{u}) - V_{\mathcal{C}}^{\phi_1}(\bm{x})$, \mbox{$A^{\phi_2}_{\bm{\zeta}}(\bm{x},\bm{u}) := Q_{\bm{\zeta}}(\bm{x},\bm{u}) - V_{\bm{\zeta}}^{\phi_2}(\bm{x})$}.

\begin{algorithm}
\caption{SAMBA: Safe Model-Based \& Active Reinforcement Learning}
\label{Algo:SAMBA_detailed}
\begin{algorithmic}[1]
\STATE \textbf{Inputs:} Risk-level $\alpha$, safety threshold $\xi$, 
\STATE \textbf{Hyper-parameters:} Data batch size $B$, model traces batch size $N$, \#\text{env-iterations} $J$, \#\text{control-iterations} $K$
\STATE \textbf{Initialisation:} GP hyper-parameters (\textsection~\ref{Sec:MBRL}), $\mathcal{D}=\emptyset$, $\lambda_{\text{CVaR}} \geq 0$ arbitrarily,  policy parameter $\bm{\theta}^{[0][0]}$ and critic parameters $\phi_{1}^{[0][0]}$, $\phi_{2}^{[0][0]}$ randomly
\STATE \textbf{for} $j=0:J-1$ \textbf{do:}
\STATE \hspace{1em} Sample batch of $B$ new traces $\mathcal{N}_j=\{\bm{\tau}_\ell\}_{\ell=1}^{\ell=B}$ from the real environment using $\bm{\theta}^{[j][0]}$,

\STATE \hspace{1em} \underline{Transform the traces $\mathcal{N}_j$ and append to data set $\mathcal{D}$:} 
\STATE \hspace{2em} \textbf{for each}  $\bm{\tau}_\ell \in \mathcal{N}_j$ \textbf{do}
\STATE \hspace{3em} \textbf{for each}  $(\bm{x}_t,\bm{u}_t,\bm{x}_{t+1}) \in  \bm{\tau}_\ell$ \textbf{do}
\STATE \hspace{4em} $\mathcal{D}=\mathcal{D} \cup \{\langle\tilde{\bm{x}}^{(\ell,t)}=(\bm{x}_t,\bm{u}_t),\bm{y}^{(\ell,t)}=\bm{x}_{t+1} \rangle\}$ (\textsection~\ref{Sec:SafeExp})

\STATE \hspace{1em} Update $\mathcal{P}_{\text{GP}}^{[j]}(\cdot)$ using $\mathcal{D}$ and GP equations (\textsection~\ref{Sec:MBRL})

\STATE \hspace{1em} \textbf{for} $k=0:K-1$ \textbf{do:}
\STATE \hspace{2em} Sample $N$ traces $\mathcal{S}=\{\bm{\tau}_\kappa\}_{\kappa=1}^N$ from $\mathcal{P}^{[j]}_{\text{GP}}(\cdot)$ using $\bm{\theta}^{[j][k]}$

\STATE \hspace{2em} \underline{Compute estimates of $\nabla_{\bm{\theta}} \mathbb{E}_{\bm{\tau}}[\mathcal{C}(\bm{\tau})]$ and $\nabla_{\bm{\theta}} \mathbb{E}_{\bm{\tau}}[\bm{\zeta}_{\text{LOO}}(\bm{\tau})]$ using $\mathcal{S}$:}

\STATE \hspace{3em} $\nabla_{\bm{\theta}} \mathbb{E}_{\bm{\tau}}[\mathcal{C}(\bm{\tau})] \approx \frac{1}{N}\displaystyle\sum_{\bm{\tau}_\kappa \in \mathcal{S}}\displaystyle\sum_{t =0}^T \nabla_{\bm{\theta}}\log \pi_{\bm{\theta}}(\bm{u}^{(\kappa)}_{t}|\bm{x}^{(\kappa)}_{t})A_{\mathcal{C}}^{\phi^{[j][k]}_1}(\bm{x}^{(\kappa)}_{t},\bm{u}^{(\kappa)}_{t})$
\STATE \hspace{3em} $\nabla_{\bm{\theta}} \mathbb{E}_{\bm{\tau}}[\bm{\zeta}_{\text{LOO}}(\bm{\tau})]\approx \frac{1}{N}\displaystyle\sum_{\bm{\tau}_\kappa \in \mathcal{S}}\displaystyle\sum_{t =0}^T \nabla_{\bm{\theta}}\log \pi_{\bm{\theta}}(\bm{u}^{(\kappa)}_{t}|\bm{x}^{(\kappa)}_{t})A_{\bm{\zeta}}^{\phi_2^{[j][k]}}(\bm{x}^{(\kappa)}_{t},\bm{u}^{(\kappa)}_{t})$
\STATE \hspace{2em} \underline{Solve for \small $\lambda_{\pi}^{[j][k],\star} = \arg\min_{\lambda_{\pi}} || \lambda_{\pi} \nabla_{\bm{\theta}} \mathbb{E}_{\bm{\tau}}[\mathcal{C}(\bm{\tau})] - (1 - \lambda_{\pi})\nabla_{\bm{\theta}} \mathbb{E}_{\bm{\tau}}[\bm{\zeta}_{\text{LOO}}(\bm{\tau})]||_{2}^{2}$:}
\STATE \hspace{3em} \textbf{if }  $\nabla_{\bm{\theta}} \mathbb{E}_{\bm{\tau}}[\mathcal{C}(\bm{\tau})]^{\mathsf{T}}\nabla_{\bm{\theta}} \mathbb{E}_{\bm{\tau}}[\bm{\zeta}_{\text{LOO}}(\bm{\tau})] \geq \|\nabla_{\bm{\theta}} \mathbb{E}_{\bm{\tau}}[\mathcal{C}(\bm{\tau})]\|_2^2$ 
\textbf{ then } $\lambda_{\pi}^{[j][k],\star} =1$
\STATE \hspace{3em} \textbf{else if } $\nabla_{\bm{\theta}} \mathbb{E}_{\bm{\tau}}[\mathcal{C}(\bm{\tau})]^{\mathsf{T}}\nabla_{\bm{\theta}} \mathbb{E}_{\bm{\tau}}[\bm{\zeta}_{\text{LOO}}(\bm{\tau})] \geq \| \nabla_{\bm{\theta}} \mathbb{E}_{\bm{\tau}}[\bm{\zeta}_{\text{LOO}}(\bm{\tau})]\|_2^2$ \textbf{ then } $\lambda_{\pi}^{[j][k],\star} =0$
\STATE \hspace{3em} \textbf{else}
\STATE \hspace{4em} $\lambda_{\pi}^{[j][k],\star} = \frac{(\nabla_{\bm{\theta}} \mathbb{E}_{\bm{\tau}}[\bm{\zeta}_{\text{LOO}}(\bm{\tau})]-\nabla_{\bm{\theta}} \mathbb{E}_{\bm{\tau}}[\mathcal{C}(\bm{\tau})])^{\mathsf{T}}\nabla_{\bm{\theta}} \mathbb{E}_{\bm{\tau}}[\bm{\zeta}_{\text{LOO}}(\bm{\tau})]}{\| \nabla_{\bm{\theta}} \mathbb{E}_{\bm{\tau}}[\bm{\zeta}_{\text{LOO}}(\bm{\tau})]-\nabla_{\bm{\theta}} \mathbb{E}_{\bm{\tau}}[\mathcal{C}(\bm{\tau})]\|^2_2}$

\STATE \hspace{2em} \underline{Update $\phi^{[j][k]}_{1}$}
\vspace{-3mm}
\begin{align*}
\hspace{3em} \phi^{[j][k+1]}_{1} = \phi^{[j][k]}_{1} - \frac{\eta^{\phi_1}_k}{NT}\sum_{\bm{\tau}_\kappa \in \mathcal{S}}\sum_{t=0}^T\left(V_{\mathcal{C}}^{\phi_1^{[j][k]}}(\bm{x}^{(\kappa)}_t)-\sum_{t' = t}^T\gamma^{t'-t}c(\bm{x}_{t'}^{(\kappa)},\bm{u}_{t'}^{(\kappa)})\right)\nabla_{\phi_1}V_{\mathcal{C}}^{\phi_1^{[j][k]}}(\bm{x}^{(\kappa)}_t)
\end{align*}
\STATE \vspace{-5mm}\hspace{2em} \underline{Update $\phi^{[j][k]}_{2}$}
\vspace{-3mm}
\begin{align*}
\hspace{3em} \phi^{[j][k+1]}_{2} = \phi^{[j][k]}_{2} - \frac{\eta^{\phi_2}_k}{NT}\sum_{\bm{\tau}_\kappa \in \mathcal{S}}\sum_{t=0}^T\left(V_{\bm{\zeta}}^{\phi_2^{[j][k]}}(\bm{x}^{(\kappa)}_t)-\sum_{t' = t}^T\gamma^{t'-t}\zeta_{\text{LOO}} (\bm{x}_{t'}^{(\kappa)},\bm{u}_{t'}^{(\kappa)})\right)\nabla_{\phi_1}V_{\bm{\zeta}}^{\phi_2^{[j][k]}}(\bm{x}^{(\kappa)}_t)
\end{align*}
\STATE \vspace{-5mm}\hspace{2em} \underline{Update $\bm{\theta}^{[j][k]}$ (\textsection~\ref{Sec:SolMethod})}
\vspace{-3mm} 
\begin{align*}
\hspace{3em} \bm{\theta}^{[j][k+1]} = \bm{\theta}^{[j][k]} - \frac{\eta^{\bm{\theta}}_{k}}{N}\displaystyle\sum_{\bm{\tau}_\kappa \in \mathcal{S}}\Big[ &\sum_{t = 0}^T \nabla_{\bm{\theta}}\log \pi_{\bm{\theta}}(\bm{u}^{(\kappa)}_{t}|\bm{x}^{(\kappa)}_{t})\Big(\lambda_{\pi}^{\star}A_{\mathcal{C}}^{\phi^{[j][k]}_1}(\bm{x}^{(\kappa)}_{t},\bm{u}^{(\kappa)}_{t})\\      &\qquad- (1 -\lambda_{\pi}^{\star})A_{\bm{\zeta}}^{\phi_2^{[j][k]}}(\bm{x}^{(\kappa)}_{t},\bm{u}^{(\kappa)}_{t})\Big)+ \lambda_{\text{CVaR}}\nabla_{\bm{\theta}}\text{CVaR}_{\alpha}(\mathcal{L}(\bm{\tau}_\kappa))\Big]
\end{align*}
\STATE \hspace{1em} Set \mbox{$\bm{\theta}^{[j+1][0]}=\bm{\theta}^{[j][K]}$}, \mbox{$\phi_{1}^{[j+1][0]}=\phi_{1}^{[j][K]}$}, \mbox{$\phi_{2}^{[j+1][0]}=\phi_{2}^{[j][K]}$}
\STATE \textbf{Return } $ \bm{\theta}^{[J][0]}$
\end{algorithmic}
\end{algorithm}

\section{Active metrics}
\subsection{Active \mbox{(semi-)metrics} intuition}
We implement our code in GPyTorch~\cite{gardner2018gpytorch}, which computes a Cholesky decomposition of $\bm{A} = \bm{L}\bm{L}^{\mathsf{T}}$ during training. 
The use of the triangular matrix $\bm{L}$ allows the problem to be divided in such a way that memory requirements are reduced, as most matrix multiplications can be pre-computed and re-used when needed.
Note also that the diagonal elements of $\bm{A}$ can be readily computed as $a_{i,i} = \sum_{j=1}^{i} L_{i,j}^2$. All these operations considered, the total computational burden of a full-set LOO distribution is reduced from $\mathcal{O}(\text{N}^4)$ to $\mathcal{O}(\text{N}^3)$. The space complexity of our batch operations, however, can increase due to the need of storing a large tensor. Rather than doing so, we implement a batched version that eliminates such a need and achieves desirable results.  

To illustrate what information the LOO semi-metric brings, we analyse under which conditions its value is null. 
Coming back to our definition, we can write explicitly the KL-divergence from $p(\bm{f}_{\star} | \mathcal{D}_{\neg i})$ to $p(\bm{f}_{\star} | \mathcal{D})$, for a single test data-point (i.e., $n_2 = 1$):
\begin{equation}
    \textrm{KL}\left(p(\bm{f}_{\star} | \mathcal{D}_{\neg i}) \| p(\bm{f}_{\star} | \mathcal{D})\right) = \frac{1}{2}\left( \underbrace{\left(\frac{\mu_{\star}-\mu_{\star}^{(i)}}{\sigma_{{n_{\text{2}}}, {n_{\text{2}}}}}\right)^2}_{\geq 0}+\underbrace{\left(\frac{{\sigma}_{{n_{\text{2}}, {n_{\text{2}}}}}^{(i)}}{{\sigma}_{{n_{\text{2}}}, {n_{\text{2}}}}}\right)^2 -  \log\left({\frac{{{\sigma}_{{n_{\text{2}}}, {n_{\text{2}}}}}^{(i)}}{\sigma_{{n_{\text{2}}}, {n_{\text{2}}}}}}\right)^2-1}_{\geq 0}\right), \label{KL-div-LOO}
\end{equation}
where we use non-bold and non-capitalised fonts in order to signify that $\sigma_{n_2, n_2}^{(i)}$, $\sigma_{n_2, n_2}$, $\mu_{\star}^{(i)}$ and $\mu_{\star}$ are scalars. Note that the second expression is non-negative since $\log(x) \le x - 1$ for all $x$ and the equality is achieved if and only if $x = 1$. Therefore for the KL divergence in Equation~\ref{KL-div-LOO} to be zero, two main conditions need to be met: $\mu_{\star}^{(i)}=\mu_{\star}$ and $\sigma_{n_2, n_2}^{(i)} = \sigma_{n_2, n_2}$. Since in general $\sigma_{n_2, n_2}^{(i)} = \sigma_{n_2, n_2} +\bm{K}_{{n_{\text{2}}},{n_{\text{1}}}} \bm{a}_{i}^{\mathsf{T}}\bm{a}_{i}\bm{K}_{{n_{\text{2}}},{n_{\text{1}}}} / a_{i, i}$ we obtain another necessary condition for the KL divergence being equal to zero: $\bm{K}_{{n_{\text{2}}},{n_{\text{1}}}} \bm{a}_{i}^{\mathsf{T}} = 0$. Now we will show that $\bm{K}_{{n_{\text{2}}},{n_{\text{1}}}} \bm{a}_{i}^{\mathsf{T}} = 0$ (and hence $\sigma_{n_2, n_2}^{(i)} = \sigma_{n_2, n_2}$) is also sufficient. Developing the condition $\mu_{\star}^{(i)} =\mu_{\star}$ we have $\bm{K}_{{n_{\text{2}}},{n_{\text{1}}}} \bm{a}_{i}^{\mathsf{T}} \bm{a}_{i} \bm{y}_{n_{\text{1}}} = 0$ since $\mu_{\star}^{(i)}=\mu_{\star}-\bm{K}_{{n_{\text{2}}},{n_{\text{1}}}} \bm{a}_{i}^{\mathsf{T}} \bm{a}_{i} \bm{y}_{n_{\text{1}}}$. As both $\bm{K}_{{n_{\text{2}}},{n_{\text{1}}}} \bm{a}_{i}^{\mathsf{T}}$ and $\bm{a}_{i} \bm{y}_{n_{\text{1}}}$ are scalars this equivalently amounts to $\bm{K}_{{n_{\text{2}}},{n_{\text{1}}}} \bm{a}_{i}^{\mathsf{T}} = 0$ or $\bm{a}_{i} \bm{y}_{n_{\text{1}}}=0$. Therefore if $\bm{K}_{{n_{\text{2}}},{n_{\text{1}}}} \bm{a}_{i}^{\mathsf{T}} = 0$ (or equivalently $\sigma_{n_2, n_2}^{(i)} = \sigma_{n_2, n_2}$) then $\mu_{\star} =\mu_{\star}^{(i)}$ and the KL divergence in Equation~\ref{KL-div-LOO} is equal to zero. 
In other words, the KL divergence between the two distribution will be zero if and only if the covariance of the test data and the target training data is zero when all other points are observed. As a consequence, the LOO semi-metric can be seen as a proxy to the expected reduction of uncertainty about training output values conditioned on the observation of the queried datapoint. 

The intuition behind these conditions is that the expected KL divergence between LOO and the full distributions answers a completely different question when compared to entropy bonus: whereas entropy bonus indicates how much ``absolute'' uncertainty there is about a \textit{test} output -- which we always expect to be high for points situated at a distance of the training inputs -- the LOO semi-metric indicates how much, on average, we can learn (i.e. how big would be the reduction in uncertainty would be) about the \textit{training} outputs if we were to observe the queried test data. Therefore, the LOO semi-metric truly answers the question of how much we can learn about the current model when observing the queried point.



\subsection{LOO computation}
Without loss of generality, assume that we leave the last sample out, i.e., we will set $i = n_1$. We have the following expressions for the means:
\begin{equation}
    \begin{aligned}
        \bm{\mu}_{\star} & = \bm{K}_{n_2, n_1}\bm{A}_{n_1, n_1} \bm{y}_{n_1}, \\
        \bm{\mu}_{\star}^{(n_1)} & = \bm{K}_{n_2, n_1-1}\bm{A}_{n_1-1, n_1-1} \bm{y}_{n_1-1},
    \end{aligned}
\end{equation}
and the covariance matrices:
\begin{equation}
    \begin{aligned}
        \bm{\Sigma}_{n_2, n_2} & = \bm{K}_{n_2, n_2} - \bm{K}_{n_2, n_1}\bm{A}_{n_1, n_1} \bm{K}_{n_1, n_2}, \\
        \bm{\Sigma}_{n_2, n_2}^{(n_1)} & = \bm{K}_{n_2, n_2} - \bm{K}_{n_2, n_1-1}\bm{A}_{n_1-1, n_1-1} \bm{K}_{n_1-1, n_2}, 
    \end{aligned}
\end{equation}
where 
\begin{equation*}
\begin{aligned}
    \bm{A}_{n_1, n_1} &= \left[\bm{K}_{n_1, n_1}+\sigma_{\omega}^{2}\bm{I}\right]^{-1}, \\
    \bm{A}_{n_1-1, n_1-1} &= \left[\bm{K}_{n_1-1, n_1-1}+\sigma_{\omega}^{2}\bm{I}\right]^{-1}.
\end{aligned}
\end{equation*}
The main difficulty in computing the updates $\bm{\mu}_{\star}^{(n_1)} -  \bm{\mu}_{\star}$, $\bm{\Sigma}_{n_2, n_2}^{(n_1)} -  \bm{\Sigma}_{n_2, n_2}$ is actually dealing with the update involving the matrices $\bm{A}_{n_1-1, n_1-1}$ and $\bm{A}_{n_1, n_1}$. 

Note that there exist $\bm{b}_0$, $b_1$, $c_0$ such that:
\begin{equation}
    \begin{aligned}
        \bm{K}_{n_2, n_1} = \begin{bmatrix}
        \bm{K}_{n_2, n_1-1} & c_0
        \end{bmatrix} \\
        \bm{K}_{n_1, n_1} = \begin{bmatrix}
        \bm{K}_{n_1-1, n_1-1} & \bm{b}_0^{\mathsf{T}} \\
        \bm{b}_0 & b_1
        \end{bmatrix}
    \end{aligned}
\end{equation}

We have 
\begin{equation*}
    \bm{A}_{n_1, n_1} = \left[\bm{K}_{n_1, n_1}+\sigma_{\omega}^{2}\bm{I}\right]^{-1}  = \begin{bmatrix}
    \bm{A}_{\neg n_1,\neg n_1} & \bm{a}_{n_1, \neg n_1}^{\mathsf{T}} \\ \bm{a}_{n_1,\neg n_1} & a_{n_1,n_1}
    \end{bmatrix}
\end{equation*}

As $\bm{A}_{n_1, n_1}^{-1}= \bm{K}_{n_1, n_1}+\sigma_{\omega}^{2}\bm{I}$, 
using the block inversion lemma it is straightforward to show that:
\begin{equation*}
        \bm{K}_{n_1-1, n_1-1}+\sigma_{\omega}^{2}\bm{I} =
    \left[\bm{A}_{\neg n_1,\neg n_1} -\bm{a}_{n_1, \neg n_1}^{\mathsf{T}} a_{n_1,n_1}^{-1} \bm{a}_{n_1,\neg n_1} \right]^{-1},\end{equation*}
and consequently:    
\begin{equation*}
    \bm{A}_{n_1-1,n_1 -1} =     \left[\bm{K}_{n_1-1, n_1-1}+\sigma_{\omega}^{2}\bm{I}\right]^{-1} =
    \bm{A}_{\neg n_1,\neg n_1} -\bm{a}_{n_1, \neg n_1}^{\mathsf{T}} a_{n_1,n_1}^{-1} \bm{a}_{n_1,\neg n_1},
\end{equation*}
For simplicity, let us introduce the following matrix:
\begin{multline*}
\bm{\Delta} = \bm{A}_{n_1, n_1} -  \begin{bmatrix} \bm{A}_{n_1-1, n_1-1}  & 0 \\
        0 & 0
        \end{bmatrix}  = \\ \begin{bmatrix}\bm{A}_{\neg n_1,\neg n_1} & \bm{a}_{n_1,\neg n_1}^{\mathsf{T}} \\ \bm{a}_{ n_1,\neg n_1} & a_{n_1,n_1} \end{bmatrix}- \begin{bmatrix} \bm{A}_{\neg n_1,\neg n_1}- \bm{a}_{n_1, \neg n_1}^{\mathsf{T}} a_{n_1,n_1}^{-1} \bm{a}_{n_1,\neg n_1}  & 0 \\0 & 0\end{bmatrix}  = \\
\begin{bmatrix} \bm{a}_{n_1, \neg n_1}^{\mathsf{T}} a_{n_1,n_1}^{-1} \bm{a}_{n_1,\neg n_1} & \bm{a}_{n_1,\neg n_1}^{\mathsf{T}} \\ \bm{a}_{ n_1,\neg n_1} & a_{n_1,n_1} \end{bmatrix} =
\begin{bmatrix} \bm{a}_{ n_1, \neg n_1}^{\mathsf{T}} \\ a_{n_1, n_1} \end{bmatrix}
\begin{bmatrix} \bm{a}_{ n_1, \neg n_1} & a_{n_1, n_1}  \end{bmatrix}/a_{n_1,n_1} =
\frac{\bm{a}_{n_1}^{\mathsf{T}} \bm{a}_{n_1}}{a_{n_1,n_1}},
\end{multline*}
where $\bm{a}_{n_1} =\begin{bmatrix} \bm{a}_{ n_1, \neg n_1} & a_{n_1, n_1}  \end{bmatrix}$ is the $n_1$-th row of the matrix $\bm{A}_{n_1, n_1}$.

Now: 
\begin{multline*}
 \bm{\Sigma}_{n_2, n_2}^{(n_1)} - \bm{\Sigma}_{n_2, n_2} =
     \bm{K}_{n_2, n_1} \bm{A}_{n_1, n_1} \bm{K}_{n_1, n_2} - 
     \bm{K}_{n_2, n_1-1}\bm{A}_{n_1-1, n_1-1} \bm{K}_{n_1-1, n_2} = \\
     \begin{bmatrix}
     \bm{K}_{n_2, n_1-1} & c_0
        \end{bmatrix}\bm{\Delta}
        \begin{bmatrix}
        \bm{K}_{n_1-1, n_2} \\ c_0
        \end{bmatrix} = 
        \bm{K}_{n_2, n_1} \frac{\bm{a}_{n_1}^{\mathsf{T}} \bm{a}_{n_1}}{a_{n_1,n_1}} \bm{K}_{n_1, n_2} .
\end{multline*}
Similarly, 
\begin{multline*}
     \bm{\mu}_{\star}^{(n_1)} - \bm{\mu}_{\star} = \bm{K}_{n_{2}, n_{1}-1} \bm{A}_{n_1-1,n_1-1} \bm{y}_{n_{1}-1} - \bm{K}_{n_{2}, n_{1}} \bm{A}_{n_1,n_1} \bm{y}_{n_{1}} = \\ -\bm{K}_{n_{2}, n_{1}} \bm{\Delta}\bm{y}_{n_{1}} = - \bm{K}_{n_{2}, n_{1}} \frac{\bm{a}_{n_1}^{\mathsf{T}} \bm{a}_{n_1}}{a_{n_1,n_1}} \bm{y}_{n_{1}}
\end{multline*}
Hence, we have that 
\begin{align*}
    \bm{\mu}_{\star}^{(i)} &= \bm{\mu}_{\star} - \bm{K}_{n_{2}, n_{1}} \frac{\bm{a}_{i}^{\mathsf{T}} \bm{a}_{i}}{a_{i,i}} \bm{y}_{n_{1}}, \\
  \bm{\Sigma}_{n_{2}, n_{2}}^{(i)} &= \bm{\Sigma}_{n_{2}, n_{2}} + \bm{K}_{n_{2}, n_{1}}\frac{\bm{a}^{\mathsf{T}}_{i} \bm{a}_{i}}{a_{i,i}} \bm{K}_{n_{1}, n_{2}}
\end{align*}
for any $i$. 

\begin{table}[h!]
    \caption{
    	Experiment hyper-parameters 
    }
    \resizebox{\columnwidth}{!}{
    \begin{tabular}{@{}llccc@{}}
    \toprule
    \multicolumn{2}{l}{} & \textbf{Safe Pendulum} & \textbf{Safe Cartpole Double Pendulum} & \textbf{Safe Fetch Reacher}         \\ \midrule
    \multicolumn{1}{c}{\multirow{13}{*}{\rotatebox{90}{Dynamics Model}}}
        & N samples                & 30                        & 30                           & 300                        \\
        & N init samples          & 30                        & 30                           & 1000                       \\
        & Optimizer                 & FullBatchLBFGS~\cite{hjmshi_2018}             & FullBatchLBFGS~\cite{hjmshi_2018}                & FullBatchLBFGS~\cite{hjmshi_2018}             \\
        & Predict delta states    & True                       & True                          & True                       \\
        & Normalize states         & True                       & True                          & True                       \\
        & GP model                 & Exact                      & Exact                         & Exact                      \\
        & Objective function       & Exact Marginal Log Likelihood & Exact Marginal Log Likelihood    & Exact Marginal Log Likelihood \\
        & kernel                    & RBF                        & RBF                           & RBF                         \\
        & CG tolerance             & 0.00001                    & 0.00001                       & 0.00001                    \\
        & Max preconditioner size & 200                        & 30                           & 150                        \\
        & Max cg iterations       & 15000                      & 15000                         & 15000                      \\
        & Optimization iterations               & 300                        & 300                           & 100                       \\
        & Learning rate            & 0.1                        & 0.1                           & 0.0001                     \\
    \midrule
    \multicolumn{1}{c}{\multirow{14}{*}{\rotatebox{90}{Agent}}} 
        & N update epochs         & 80                         & 80                            & 80                         \\
        & Mini batch size                  & 30000                      & 30000                         & 90000                      \\
        & N neurons                & 32                         & 32                            & 32                         \\
        & Policy learning rate      & 3e-4                       & 3e-4                          & 3e-4                       \\
        & Value function learning rate    & 1e-3                       & 1e-3                          & 1e-3                       \\
        & Penalty learning rate     & 5e-2                       & 5e-2                          & 5e-2                       \\
        & Clipping                  & 0.2                        & 0.2                           & 0.2                        \\
        & Max gradiant norm           & 0.5                        & 0.5                           & 1.0                        \\
        & Gamma                     & 0.99                       & 0.95                          & 0.99                       \\
        & Lambda                    & 0.97                       & 0.97                          & 0.95                       \\
        & Value function tragets    & Monte Carlo                & Monte Carlo                   & Monte Carlo                \\
        & Value function loss fn    & Squared loss               & Squared loss                  & Clipped  loss                  \\
        & CVaR cost limit           & 0.025                      &  0.025                          & 0.0166666667               \\
        & CVaR risk                 & $\alpha \in (0.5, 1.0)$                        &  $\alpha \in (0.5, 1.0)$                           & $\alpha \in (0.5, 1.0)$                        \\
                                                         \midrule
    \multicolumn{1}{c}{\multirow{3}{*}{\rotatebox{90}{Runner}}}
        & Max trajectory length     & 30                        & 30                            & 30                         \\
        & N training epochs       & 100                       & 100                           & 100                        \\
        & N epoch samples         & 30000                     & 30000                         & 90000                      \\
    \bottomrule
    \end{tabular}
    }
\end{table}

\section{Environmental settings}
We implemented a safety cost function with the following settings; The unsafe region started at  $USR_{min} = 20\pi/180$ and ended at $USR_{max} = 30\pi/180$, therefore if $\theta_1$ was in $USR_{min}~\leq~\theta_1~\leq~USR_{max}$ we would refer to this as a violation and safety cost would be incurred. The Hazard Region contained within $HZ_{min} = USR_{min} - \pi/4 \leq HZ \leq USR_{max} + \pi/4 = HZ_{max}$. Where $\theta_1$ is transformed to always remain in the region $\theta_1 \in [-\pi, \pi]$. Therefore, safety cost is linearly proportional to the distance from the edge of the hazard region ($HZ_{min}$ or $HZ_{max}$) to the centre of the hazard region $(HZ_{max}+HZ_{min})/2$. We also implemented batched versions of the pendulum reward function as well as batched versions of the pendulum safety function using torch. We also implement the unsafe region centre as the middle between the fetchers starting position and its goal (with respect to the $x$ position only), then the hazard region expands around this point with respect to $1/4$ of the total distance between the fetchers start state and goal state. Note, we terminated Safe Pendulum once the last 5 rewards in a trajectory were all $\leq -0.01$, as this provides an adequately stabilised pendulum. 

\begin{figure*}[h!]
    \begin{minipage}[t]{0.3\textwidth}
        \centering
        \includegraphics[width=\textwidth, trim={0em, 19em, 0em, 15em}, clip]{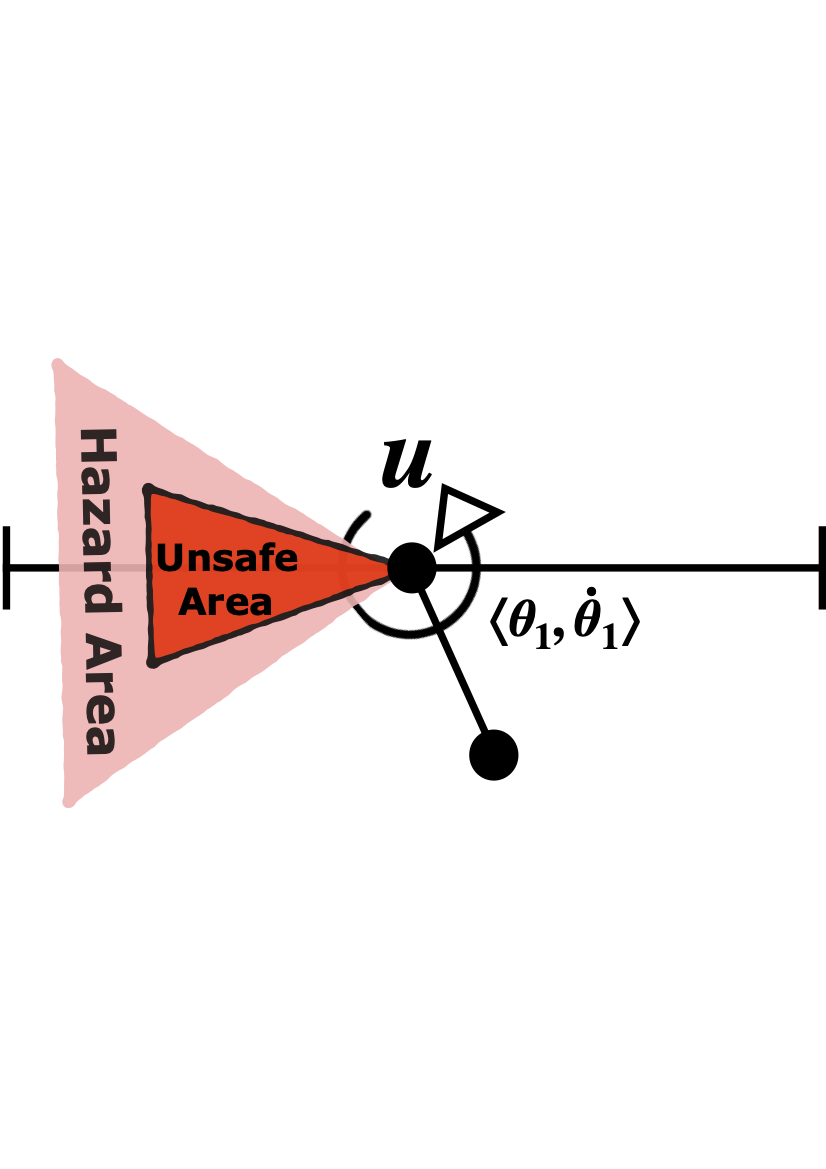}
        Safe Pendulum
    \end{minipage}
    \hfill
    \begin{minipage}[t]{0.3\textwidth}
        \centering
        \includegraphics[width=\textwidth, trim={0em, 15em, 0em, 15em}, clip]{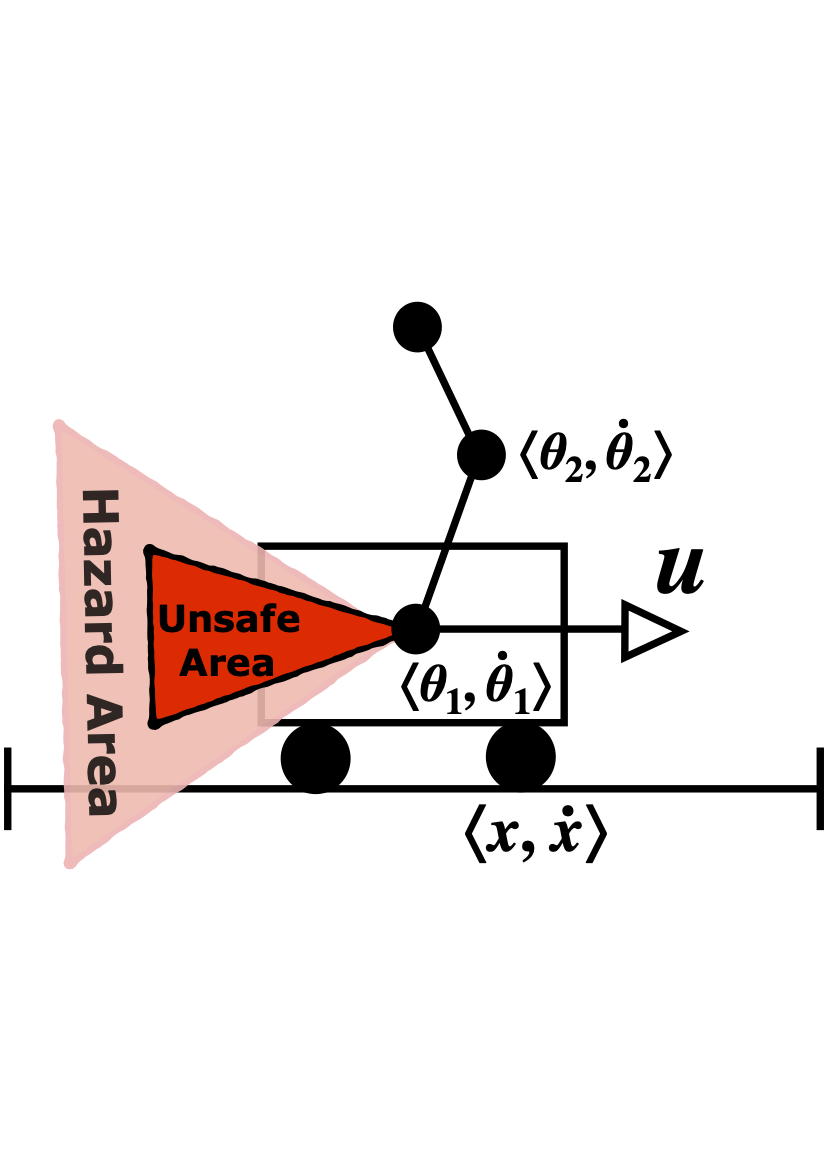}
        Safe Cartpole Double Pendulum
    \end{minipage}\hfill
    \begin{minipage}[t]{0.3\textwidth}
        \centering
        \includegraphics[width=0.9\textwidth]{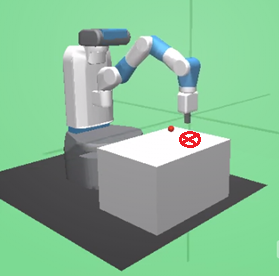}
        Safe Fetch Reacher
    \end{minipage}
    \caption{Environments with their respective unsafe regions visualized.}
    \label{fig:enviroment}
\end{figure*}
\section{Experimental Evaluation}
\subsection{LOO performance analysis}

\begin{figure*}[h!]
    \minipage{0.5\textwidth}
      \includegraphics[width=\linewidth]{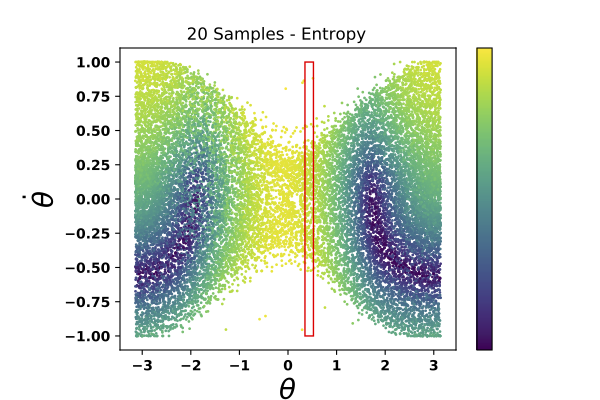}
    \endminipage\hfill
    \minipage{0.5\textwidth}
      \includegraphics[width=\linewidth]{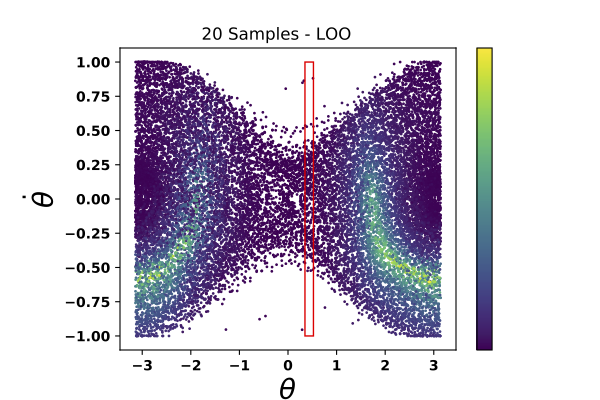}
    \endminipage\hfill
    \minipage{0.5\textwidth}
      \includegraphics[width=\linewidth]{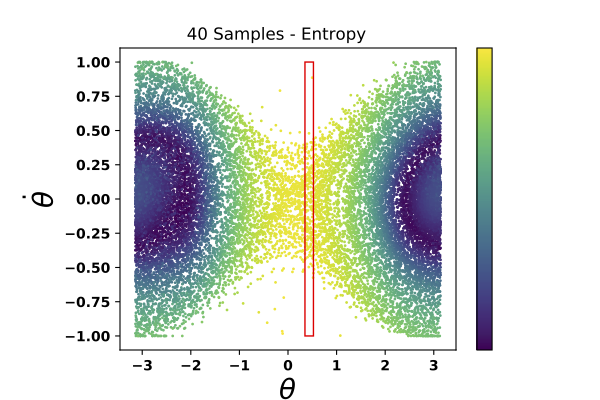}
    \endminipage\hfill
    \minipage{0.5\textwidth}
      \includegraphics[width=\linewidth]{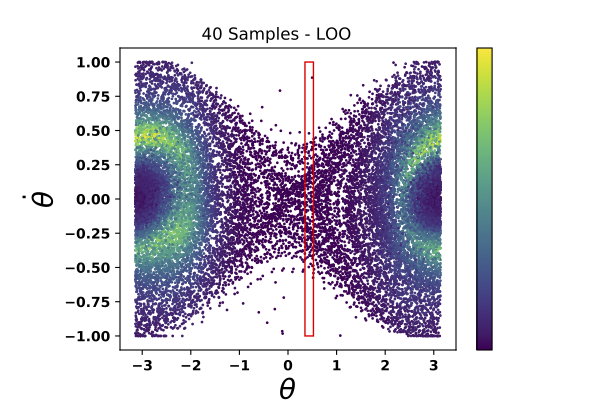}
    \endminipage\hfill
    \minipage{0.5\textwidth}
      \includegraphics[width=\linewidth]{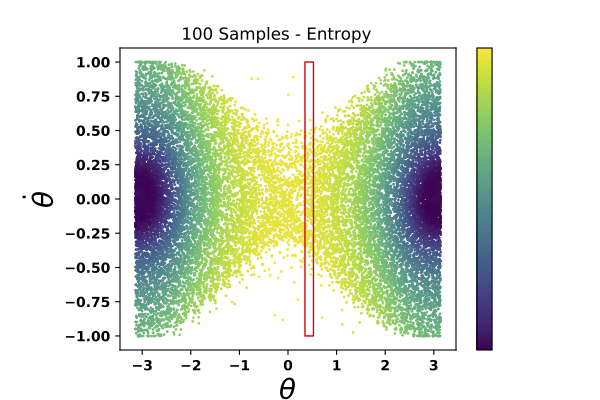}
    \endminipage\hfill
    \minipage{0.5\textwidth}
      \includegraphics[width=\linewidth]{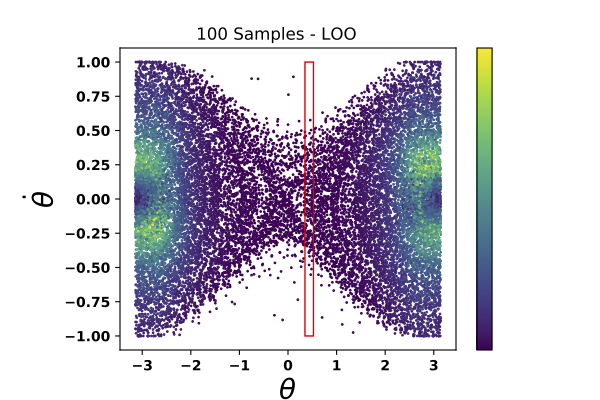}
    \endminipage
    \caption{Comparison between Entropy vs. LOO values across the entire state space of Safe Pendulum across true sampled data points. Yellow corresponds to a higher metric values, while blue to a lower. Note that PPO~(or any policy used within this framework) is encouraged to visit regions with higher~(yellow) weight. The red box indicates the unsafe region of the state space.}
    \label{fig:IG_plots}
\end{figure*}

This section empirically compares the typical entropy-based method with LOO. For comparison, we collected $N$ random rollouts from Safe Pendulum to train our GP. Next, we calculated the entropy/LOO metric for unseen random rollouts totaling $20,000$ test samples. Figure~\ref{fig:IG_plots} shows the result of this evaluation, where we compared Entropy vs. LOO  for differing sizes of training data $N$ for the GP  ($N=20$, $N=40$ and $N=100$). The red box contains the USR (Unsafe Region). The experiment suggests that LOO is much more conservative and will aim to guide the policy away from the dangerous area. At the same time, entropy encourages exploration into the unsafe area, additionally encouraged with the increase of N - a highly undesirable property when safe active exploration is required. 

\begin{figure*}[h!]
      \includegraphics[width=\linewidth, trim={0 0 0 2.2cm }, clip]{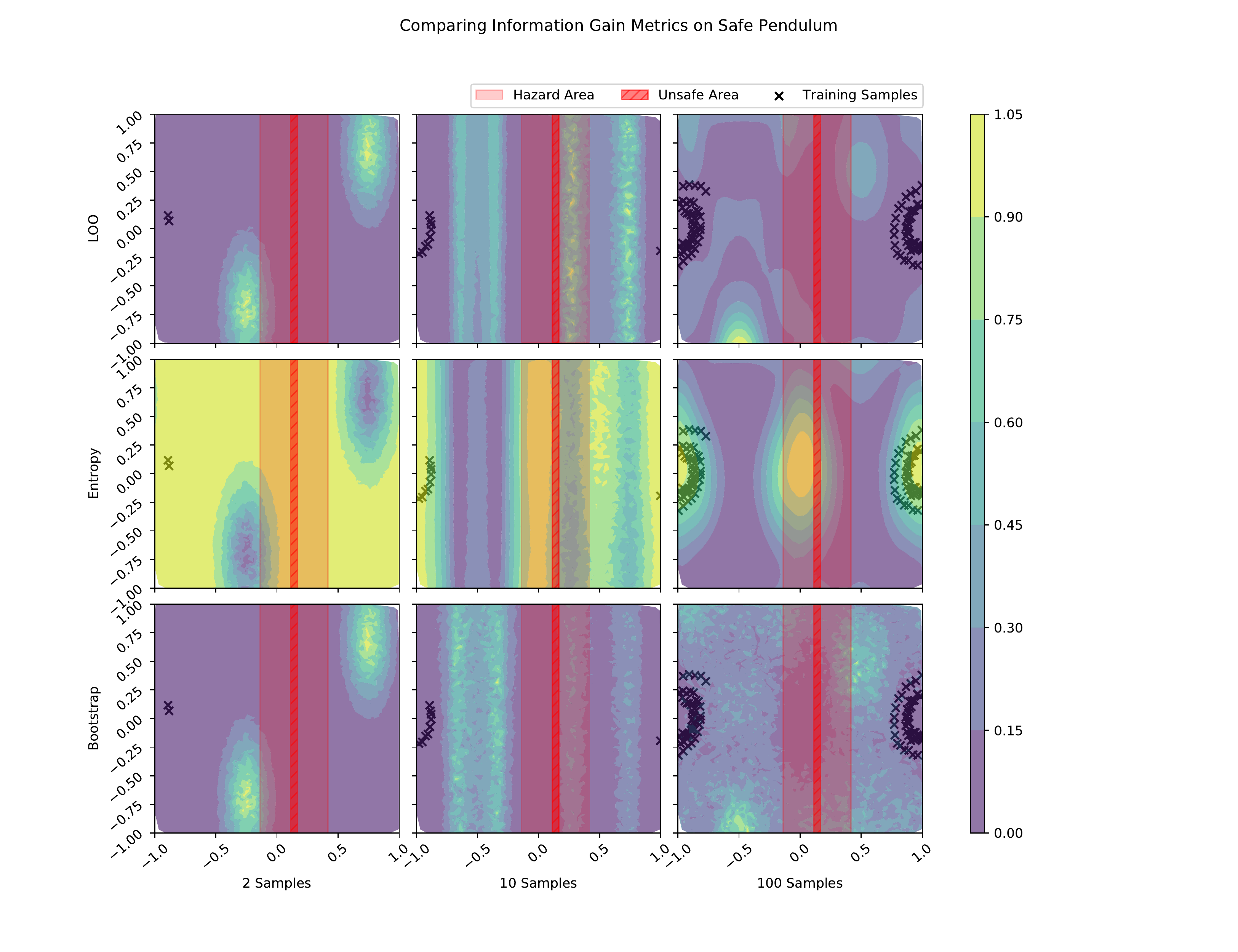}
    \includegraphics[width=\linewidth, trim={0 0 0 2.2cm }, clip]{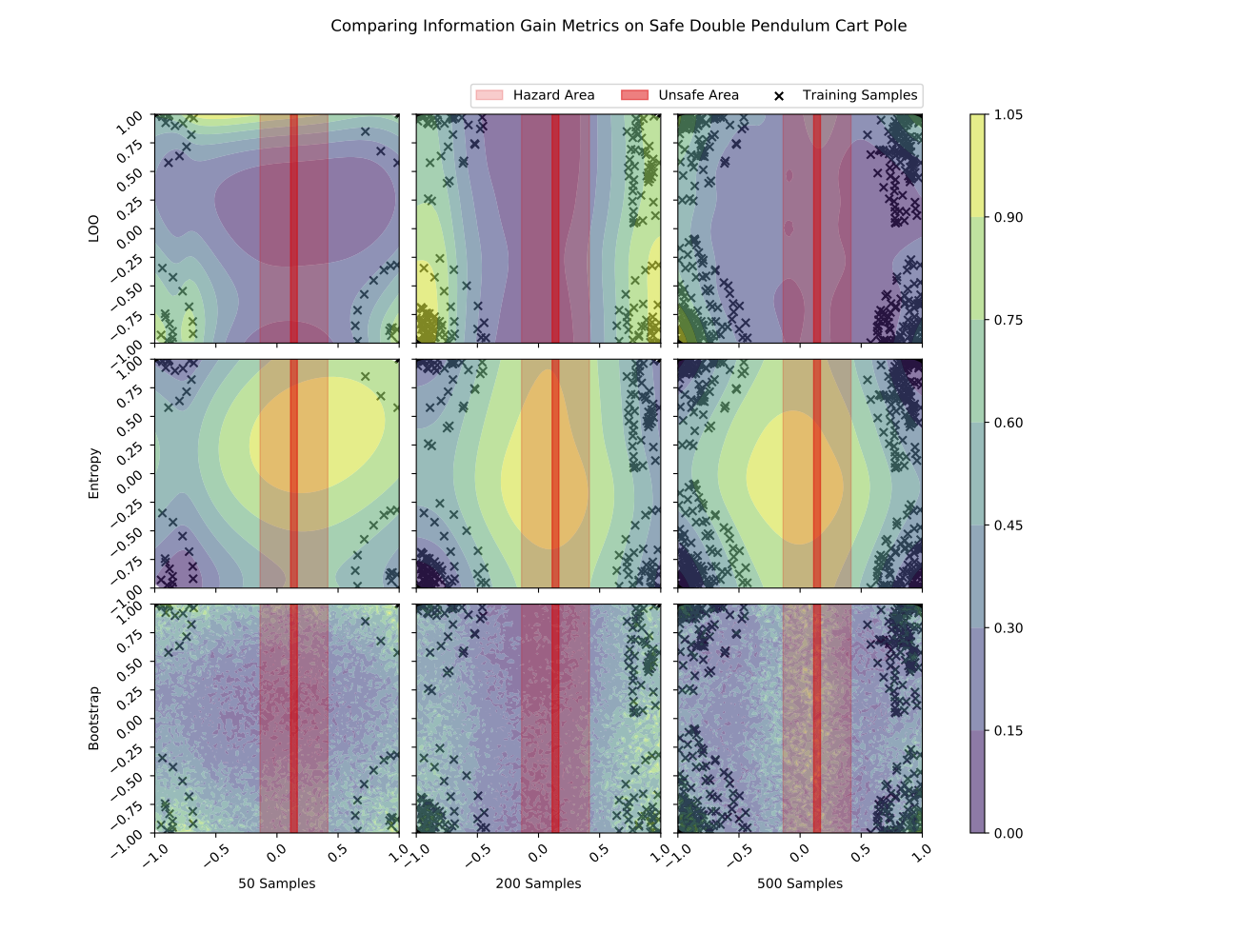}
    \caption{Comparison Entropy vs. LOO values across the entire state space of Safe Pendulum and Safe Cartpole Double Pendulum. Yellow corresponds to a higher metric value, while blue to a lower. Note that PPO (or any policy used within this framework) is encouraged to visit regions with higher (yellow) weight. The red box indicates the unsafe region of the state space.}
    \label{fig:IG_plots_fetch}
\end{figure*}


\begin{figure*}[h!]
    \minipage{\textwidth}
      \includegraphics[width=\linewidth, trim={0 0 0 2.2cm }, clip]{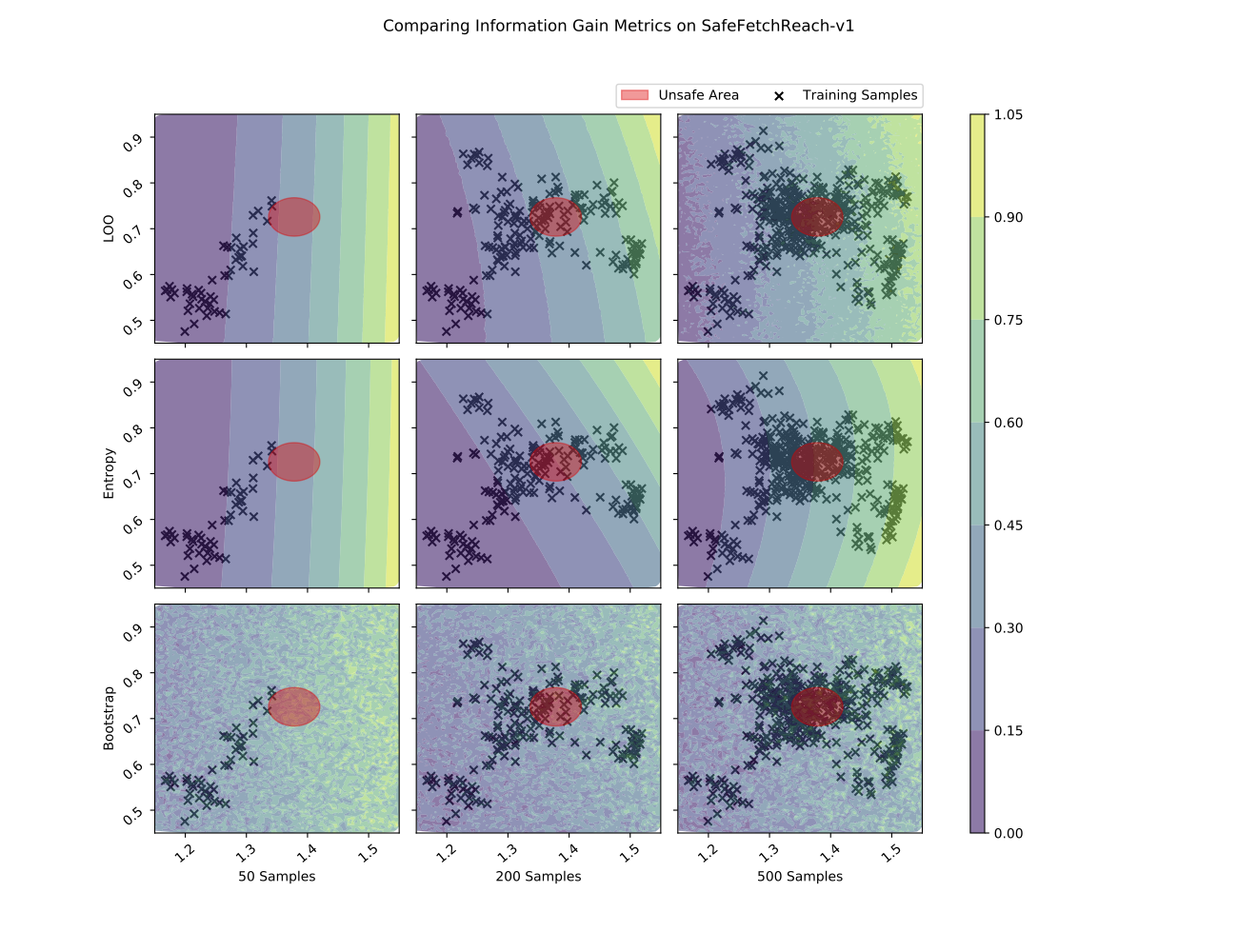}
    \endminipage
    \caption{Comparison Entropy vs. LOO  values across the entire state space of Safety Fetcher. Yellow corresponds to a higher metric value, while blue to a lower. Note that PPO (or any policy used within this framework) is encouraged to visit regions with higher (yellow) weight. The red box indicates the unsafe region of the state space.}
    \label{fig:IG_plots_fetch}
\end{figure*}

\subsection{Dynamics model analysis}

\begin{figure*}[h!]
\centering
    \minipage{1.0\textwidth}
        \centering
        \includegraphics[width=1.0\textwidth, trim={8.5em 0em 9.5em 0em}, clip]{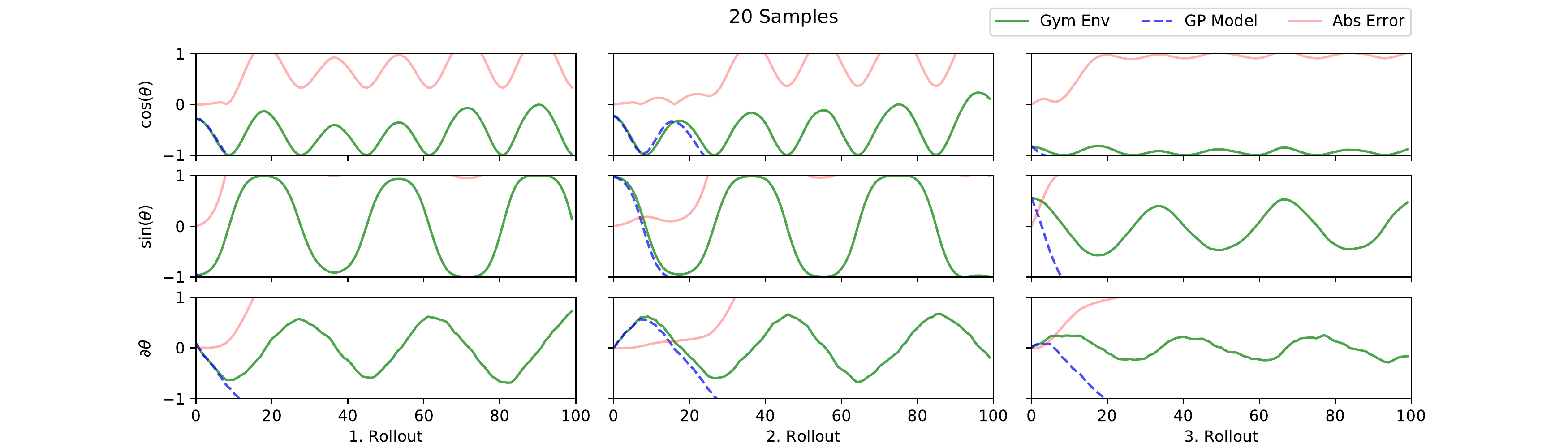}
    \endminipage \\
    \minipage{1.0\textwidth}
        \centering
        \includegraphics[width=1.0\textwidth, trim={8.5em 0em 9.5em 0em}, clip]{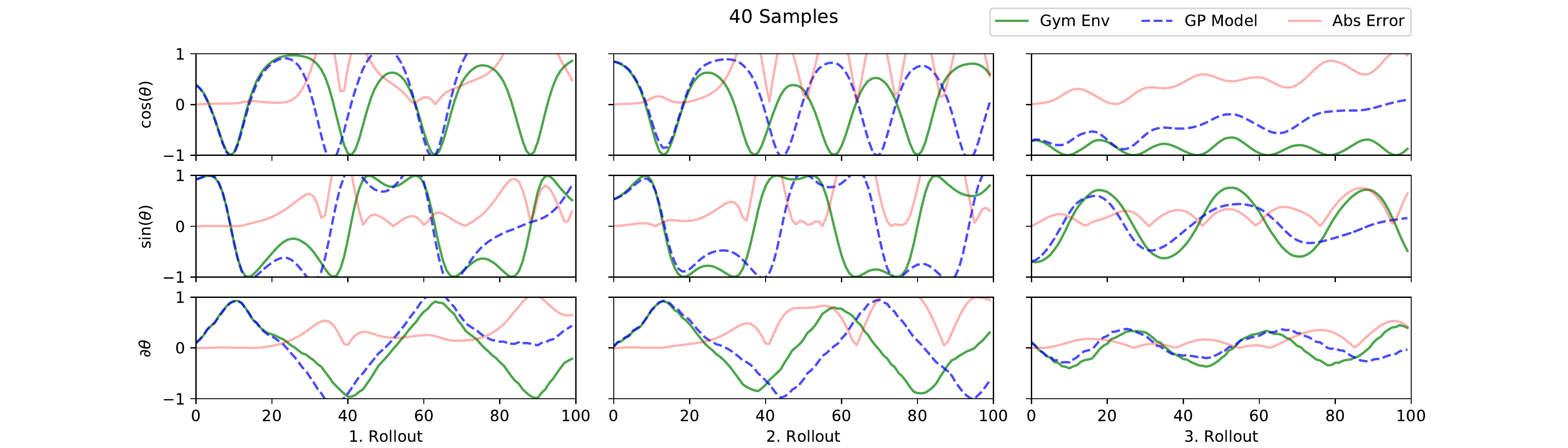}
    \endminipage \\
    \minipage{1.0\textwidth}
        \centering
        \includegraphics[width=1.0\textwidth, trim={8.5em 0em 9.5em 0em}, clip]{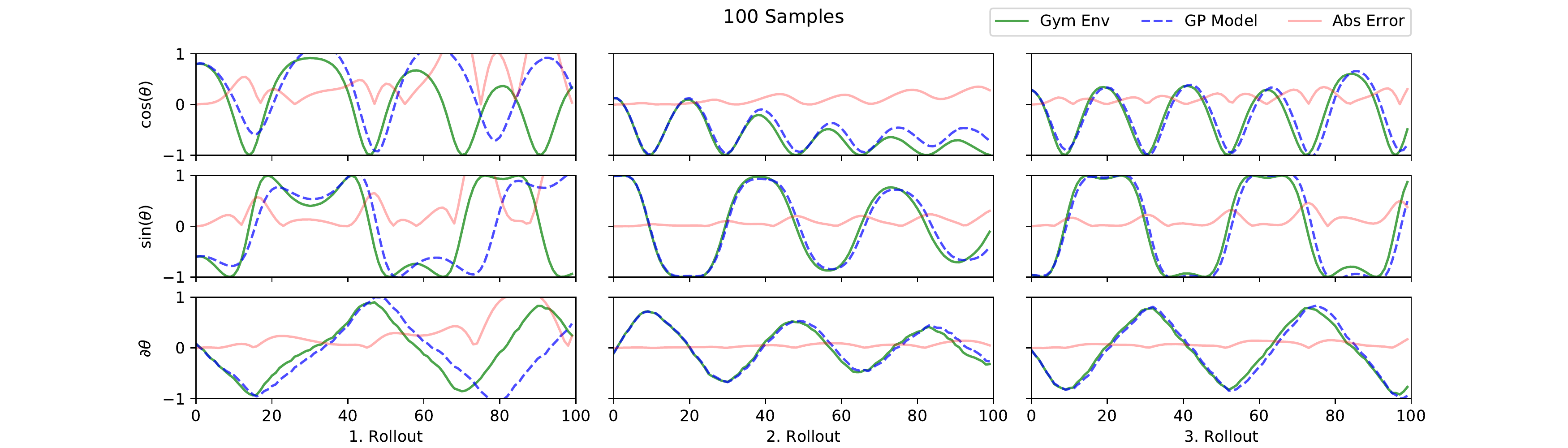}
    \endminipage
    \caption{Samples trace from Pendulum.}
     \label{fig:traces}
\end{figure*}

This section presents and compares sample traces between different GP dynamics models trained on a different number of data points (20, 40, and 100 samples). The procedure for the comparison is the following: We rollout the real environment for one full trajectory with random actions and record states and actions. For each GP, we show three sample traces of length 100 time steps (dashed blue line). The sampled traces are compared to the true trajectories by concatenating the true action taken (from the true trajectory) to the open-loop predicted states. We can see in Fig~\ref{fig:traces} how even when the dynamical system differences to the model, there is still many similarities to the transition dynamics. 

\begin{figure*}[h!]
	\centering
	\includegraphics[angle=-90, scale=0.27, trim={30em 0em 25em 0em},clip]{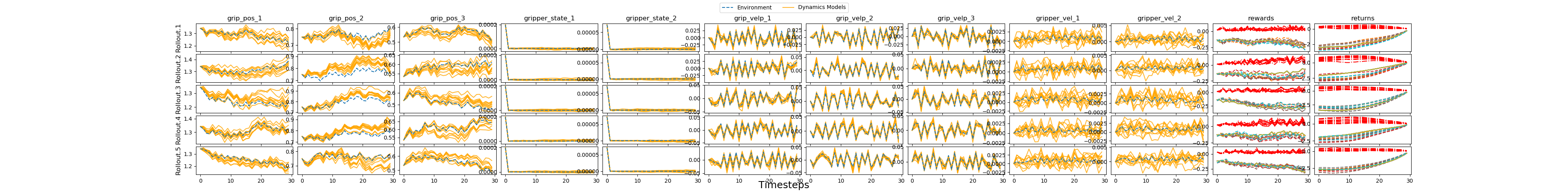}
	\caption{Sampled traces from Safe Fetch Reacher.}
	\label{fig:tracesf}
\end{figure*}

\end{document}